\pdfoutput=1

\documentclass[11pt]{article}

\usepackage{ACL2023}

\usepackage{times}
\usepackage{latexsym}

\usepackage[T1]{fontenc}

\usepackage[utf8]{inputenc}

\usepackage{microtype}

\usepackage{inconsolata}
\usepackage{hyperref}
\usepackage[utf8]{inputenc}
\usepackage{url}            
\usepackage{booktabs}       
\usepackage{amsfonts}       
\usepackage{nicefrac}       
\usepackage{microtype}      
\usepackage{xcolor}         
\usepackage{graphicx}
\usepackage{subfigure}
\usepackage{amsmath}
\usepackage{multirow}
\usepackage{algorithm, algorithmic}
\usepackage{caption}
\usepackage{pifont}
\usepackage{wrapfig,lipsum,booktabs}
\usepackage{lineno}
\usepackage{colortbl}
\definecolor{Blue2}{RGB}{235, 245, 250}

\title{Joint Generator-Ranker Learning for Natural Language Generation}

\author{Weizhou Shen\textsuperscript{\rm 1}\thanks{$\;\;$Done during his internship at Microsoft Research
Asia.}, Yeyun Gong\textsuperscript{\rm 2}\thanks{$\;\;$Corresponding authors.}, Yelong Shen\textsuperscript{\rm 3}$^\dagger$, Song Wang\textsuperscript{\rm 3}, \\
\textbf{Xiaojun Quan}\textsuperscript{\rm 4}$^\dagger$, \textbf{Nan Duan}\textsuperscript{\rm 2}, \textbf{Weizhu Chen}\textsuperscript{\rm 3}\\
\textsuperscript{\rm 1}School of Computer Science and Engineering, Sun Yat-sen University\\
\textsuperscript{\rm 2}Microsoft Research Asia, \textsuperscript{\rm 3}Microsoft Research\\
\textsuperscript{\rm 1}shenwzh3@mail2.sysu.edu.cn, \textsuperscript{\rm 2}\{yegong, nanduan\}@microsoft.com, \\
\textsuperscript{\rm 3}\{yelong.shen, sonwang, wzchen\}@microsoft.com, \textsuperscript{\rm 4}xiaojunquan@gmail.com
}

\begin{document}
\maketitle
\begin{abstract}
Generate-then-rank is a widely used mechanism for text generation, where a generator produces multiple text candidates and a ranker chooses the best one among the text candidates. However, existing methods usually train the generator and the ranker individually, neglecting the mutual feedback that could further enhance the generation quality. To tackle this limitation, we propose JGR, a novel joint training algorithm that integrates the generator and the ranker in a single framework. JGR optimizes the generator with a hybrid objective that combines data likelihood and ranker reward, and trains the ranker with a contrastive loss that compares the generator outputs. By iteratively updating the generator and the ranker, JGR can effectively harmonize their learning and enhance their quality jointly. We evaluate JGR on various text generation tasks and demonstrate that it surpasses existing methods on four public datasets across three common generation scenarios. Our code and models are publicly available at \url{https://github.com/microsoft/ProphetNet/tree/master/JGR}.
\end{abstract}

\section{Introduction}

The quality of the output texts produced by neural natural language generation (NLG) models, such as those for machine translation~\citep{vaswani2017attention} and summarization~\citep{lewis2019bart}, depends largely on how they are trained and decoded. The conventional approach is to train them with log-likelihood objectives and decode them with greedy or beam search strategies. However, this approach often fails to select the best sample with the highest evaluation score among the generated candidates, as shown by previous studies~\citep{pmlr-v97-cohen19a, meister-etal-2020-beam}.

To overcome this limitation, some recent works~\citep{liu-liu-2021-simcls, liu-etal-2021-refsum,li2022competitionlevel, ravaut2022summareranker} proposed to use a separate ranker model to re-rank the output texts of the generator model, following a generate-then-rank pipeline. This pipeline can improve the quality of the output texts by exploiting the ranker model's ability to evaluate and compare different candidates. However, this pipeline also has a drawback: it requires training the generator and ranker models in two separate phases, which may not fully exploit the generative ability of the generator model and the feedback from the ranker model.

In this paper, we propose a novel \textbf{J}oint training paradigm of both \textbf{G}enerator and \textbf{R}anker (JGR) for NLG tasks, which aims to overcome the drawback of the generate-then-rank pipeline. Unlike previous works, which train the generator and ranker models separately, we explore a joint and iterative training algorithm that updates both models in turn. Our main motivation for the joint and iterative training of the generator and ranker is twofold. First, the ranker model can provide valuable feedback to the generator model based on the ranking scores of the generated candidates. This encourages the generator model to produce better outputs. Second, the ranker model can also benefit from the outputs of a progressively better generator model, and improve its ranking performance gradually.

The JGR framework consists of a generator and a ranker. During training, the generator and ranker alternate to update their parameters, and each of them involves the other's outputs in its own input signals. Specifically, the ranker model is trained to rank the outputs generated by the generator model for a given input text by assigning a ranking score. At the generator training phase, the generator model uses a combination of the ranker score and the matching score (e.g., BLEU) as the reward for each sample, and trains with policy gradients, which encourages the generator to produce candidates with higher rewards and  mitigates the exposure bias issue in the teacher-forcing learning.

To assess the effectiveness of JGR, we conduct experiments on four diverse NLG tasks from different domains, including abstractive summarization~\citep{hermann2015cnndm}, conversational summarization~\citep{gliwa2019samsum}, question generation~\citep{rajpurkar2016squad}, and dialogue~\citep{zhang2018personalizing}. The experimental results demonstrate that JGR achieves remarkable performance gains over the conventional MLE training method, with a 3-point increase in ROUGE-2 score on the CNN/DailyMail dataset and a 3.5-point increase in BLEU-2 score on PersonaChat.

Furthermore, we make several interesting observations from the results. First, the rewards from the ranker are more effective than the rewards from the direct metrics, but combining them together stabilizes the training and produces a better generator. 
Second, training the ranker only on the candidates from the generator is better than using ground-truth as positive examples. Third, sampling more candidates during training leads to better performance within a certain range, which is consistent with data augmentation. 
Fourth, though trained with reinforcement learning aimed at optimizing automatic evaluation metrics, JGR still does not compromise on other aspects of generation quality.
Finally, the joint training paradigm increases the diversity of the generator outputs, which in turn benefits the ranker training.

\section{Related Work}
\subsection{Natural Language Generation}

Natural language generation is a long-standing research topic. RNN-based methods for dialog systems~\cite{wen2015semantically} and convolutional methods for translation~\cite{gehring2016convolutional} are some examples of earlier approaches. In the last few years, pre-trained transformer models have advanced the state of the art on many NLG tasks. These models, such as BART~\citep{lewis2019bart}, ProphetNet~\citep{qi2020prophetnet}, and T5~\citep{JMLR:v21:20-074}, use an encoder-decoder architecture and leverage large amounts of unlabeled data. Other models, such as GPT2~\citep{radford2019language} and UniLM~\citep{NEURIPS2019_c20bb2d9}, use only a decoder or an encoder for natural language generation.

Reinforcement learning can assist the training of NLG models, as shown by several works. \citet{8099614, paulus2018a} used self-critical methods that measure the reward as the difference between the metric score and the baseline score.  \citet{bahdanau2017an, le2022coderl} introduced actor-critic frameworks~\citep{konda1999actor}, which is also a joint training framework, while they have not considered the contrastive rewards between different candidates given one input. We provide a more detailed comparison in \ref{app:RL}.

Another common approach to NLG is to apply adversarial networks~\citep{NIPS2014_5ca3e9b1}. For example, SeqGAN~\citep{Yu_Zhang_Wang_Yu_2017}, RankGAN~\citep{NIPS2017_bf201d54}, GCN~\citep{lamprier2022generative} and SelfGAN~\citep{scialom2021beam}. These methods also introduce a joint training framework, however, instead of training a ranker, they trained a discriminator, which distinguishes the ground-truth text and the generator outputs.
In Appendix~\ref{app:gan}, we detail the main distinctions between these methods and our JGR.

\subsection{Generate-then-Rank Framework}

The generate-then-rank framework generates some candidate texts with a generator and then ranks them with a ranker. SimCLS~\citep{liu-liu-2021-simcls}, RefSum~\citep{liu-etal-2021-refsum}, and SumRanker~\citep{ravaut2022summareranker} train rankers separately to rank the outputs of summarization models such as BART~\citep{lewis2019bart}. In other domains, such as code generation and math problem solving, rankers are also used to evaluate the generated outputs, as shown by AlphaCode~\citep{li2022competitionlevel} and Verifier~\citep{https://doi.org/10.48550/arxiv.2110.14168}. There are also some works trying to compress the generate-then-rank pipeline to one single model using extra training objectives, for example, MATCHSUM~\citep{zhong-etal-2020-extractive}, CoLo~\citep{an-etal-2022-colo}, and BRIO~\citep{liu-etal-2022-brio} with contrastive learning, and Amortized Noisy-Channel NMT~\citep{pang2021amortized} with Q-learning. However, the above methods do not explore the joint training framework that optimizes both generators and rankers together.

\begin{figure*}[t]
    \centering
    \includegraphics[width=0.95\textwidth]{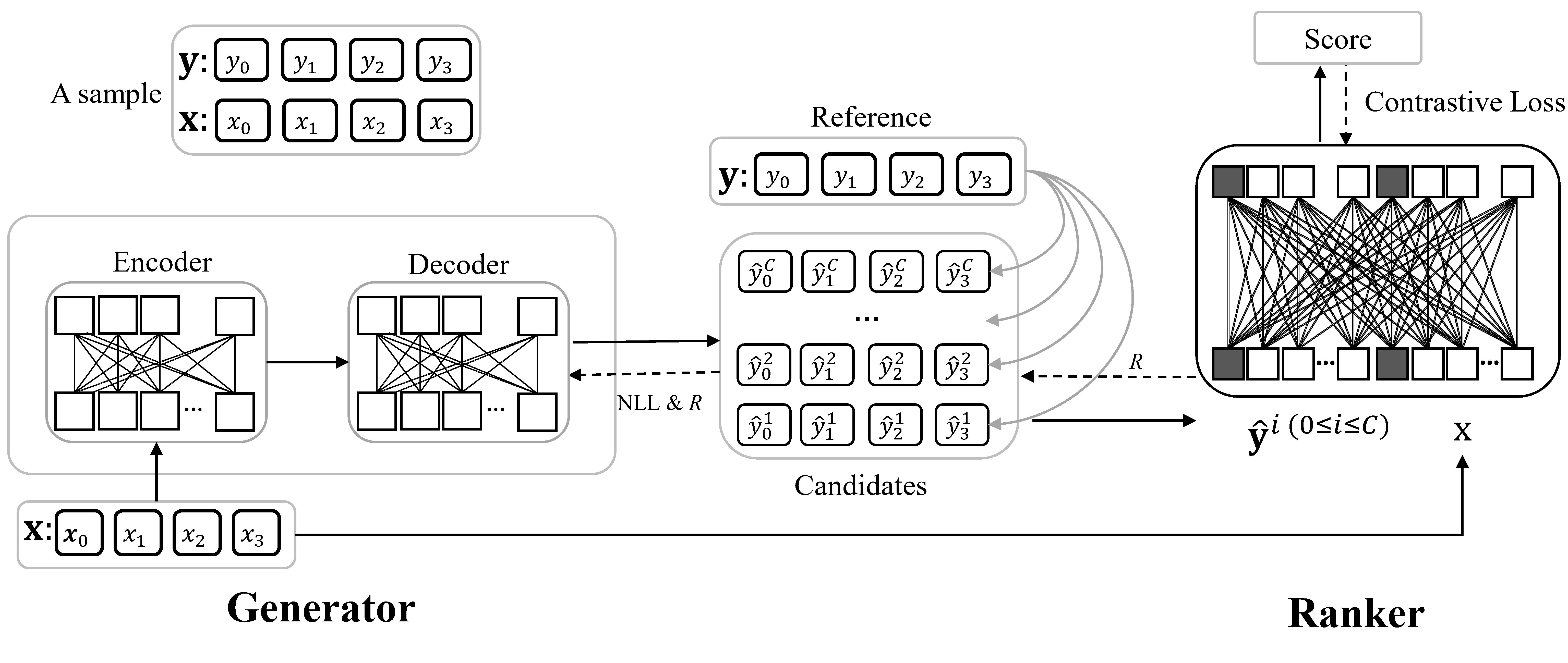}
	\caption{An example to illustrate the generator and ranker in JGR. The input text $\mathbf{x}$ is first fed into the encoder-decoder generator model to sample candidates $\mathbf{\hat{y}}^1, ..., \mathbf{\hat{y}}^C$, then the candidates are sent to ranker together with the input text to output ranker scores and feedback rewards. }
	\label{fig.jr2}
	\vspace{-0.5cm}
\end{figure*}

In the retrieve-then-rank framework for dense retrieval~\citep{karpukhin2020dense}, a retriever first finds relevant documents from a large collection, then a ranker reorders them according to their scores. Our JGR is partially motivated by this framework, we think in the generate-then-rank framework, the generation can be viewed as a retrieval process. Therefore, during training and inference, the generator should sample enough candidates for the ranker to re-rank. Several works have proposed to jointly train the retriever and the ranker to improve retrieve-then-rank framework. Such as RocketQA v2~\citep{ren2021rocketqav2} and AR2~\citep{zhang2021adversarial}. However, to our knowledge, JGR is the first work applying the joint training paradigm to the generate-then-rank framework for NLG.

\section{Methodology} \label{sec.preliminaries}
The model architecture of our JGR, shown in Figure \ref{fig.jr2}, has two components: a generator that outputs several text candidates for an input text using an encoder-decoder model, and a ranker that scores these text candidates. The JGR workflow works as follows: a) the generator generates multiple text candidates conditioned on the input text; b) the input text and the text candidates are combined and sent to the ranker; c) the ranker learns to rank the text candidates via a contrastive learning objective; d) the ranker gives a reward to each text candidate, which in turn is used to train the generator. In the following, we first introduce the basic elements of conditional text generation, including problem definition, model architecture, and model training. 

\subsection{Preliminaries}
\label{sec:basic_generator}

Given a text pair $(\mathbf{x}, \mathbf{y})$, $\mathbf{x}$ is the input text sequence, $\mathbf{y}$ is the target text sequence. The conditional text generation tasks ask the model to generate a high-quality output $\hat{\mathbf{y}}$ that close the ground-truth $\mathbf{y}$ based on the input $\mathbf{x}$. We adopt the Transformer-based~\citep{vaswani2017attention} encoder-decoder architecture as the general model for conditional text generation. The encoder part transforms $\mathbf{x}$ into a tensor representation $\mathcal{H}_e$ using the Transformer model, as shown in Eqn. ~\ref{eq:encoder}. 
\begin{equation}
\label{eq:encoder}
    \mathcal{H}_e = \textbf{Encoder}(\mathbf{x}), 
\end{equation}
The decoder part uses $\mathcal{H}_e$ as input and produces a text sequence via the auto-regressive fashion.
\begin{equation}
\label{eq:decoder}
    \mathbf{\hat{y}}  \sim {\textbf{Decoder}}(\mathbf{\hat{y}}, \mathcal{H}_e) = \prod_{t=1}^{|\hat{y}|} p(\hat{y}_t | \hat{y}_{<t}, \mathcal{H}_e). 
\end{equation}

To simplify the notation, we use $G_{\theta}(\cdot)$ to denote the encoder-decoder generation model with parameters $\theta$, and $p_{G_{\theta}}(\mathbf{\hat{y}} | \mathbf{x})$ to denote the probability of generating $\mathbf{\hat{y}}$ given $\mathbf{x}$. The standard way to train the encoder-decoder sequence generation model is to minimize the negative log-likelihood of the ground-truth target sequence:
\begin{equation}
\label{eq:LNLL}
    \mathcal{L}_\text{NLL} = - \sum_{t=1}^{|\mathbf{y}|} \log p_{G_\theta}(y_{t}|y_{<t}, \mathbf{x}).
\end{equation}

During inference of the generator, a decoding strategy such as beam search is usually adopted. However, previous studies~\citep{pmlr-v97-cohen19a, meister-etal-2020-beam} observed that the top-scored candidate from decoding strategies is often not the optimal candidate regarding the evaluation metric. Therefore, we design JGR to alleviate this problem through joint training of the generator and a ranker.

\subsection{Joint Generator-Ranker Training}\label{sec:method}

We use $G_{\theta}(\cdot)$ and  $D_{\phi}(\cdot)$ to represent the generator model and ranker model respectively, where $G_{\theta}(\cdot)$ is a text generation model with an encoder-decoder structure as explained in section ~\ref{sec:basic_generator}, and $D_{\phi}(\cdot)$ works as a scoring model that takes the concatenation of input text $\mathbf{x}$ and generated text $\mathbf{\hat{y}}$ as the input, and outputs a scalar value $s_{\hat{y}}$ reprensenting the quality of the generated text:
\begin{equation}\label{eq:ranker_score}
    s_{\hat{y}} = D_{\phi}([\mathbf{x},\mathbf{\hat{y}}])
\end{equation}

During the training stage, the generator and ranker are trained alternatively and iteratively. Algorithm \ref{algo:JGR_train} shows the training procedure of JGR. 
We first warm up the generator $G_{\theta}$ with a standard negative log-likelihood (NLL) loss according to Eqn~\ref{eq:LNLL}. Then, we iteratively update the ranker and generator as follows.

\textbf{Fix {$G_{\theta}(\cdot)$}, Train $D_{\phi}(\cdot)$}: the goal of the ranker model $D_{\phi}(\cdot)$ is to choose the best sample from a set of candidates generated by the generator model, which we denote as $\mathcal{\hat{Y}} = \{ \mathbf{\hat{y}}^1,\mathbf{\hat{y}}^2, ..., \mathbf{\hat{y}}^C \}$ 
\begin{equation}
   \{ \mathbf{\hat{y}}^1,\mathbf{\hat{y}}^2, ..., \mathbf{\hat{y}}^C \} \sim p_{G_{\theta}}( \cdot  | \mathbf{x}), 
\end{equation}
where $C$ is the number of sampled candidates. For each $\mathbf{\hat{y}}^i$, we calculate the matching score (e.g., BLEU or ROUGE) with the ground-truth text $\mathbf{y}$, denoted as $\Delta(\mathbf{y}, \mathbf{\hat{y}}^i)$. Then, we pick up the positive and negative samples in the candidate set based on $\Delta(\mathbf{y}, \mathbf{\hat{y}}^i)$ for training the ranker. Specifically, we use $\mathbf{\hat{y}}^+$ to denote the text candidate with the highest matching score, and $\mathcal{\hat{Y}}^-$, whose size is a hyper-parameter, to denote the negative candidate set containing a certain number of candidates with the lowest scores. The ranker model is trained by minimizing contrastive loss:
\begin{equation}
\label{eq:ranker_obj}
    \mathcal{L}^{\phi} = - \log p_{D_\phi}(\mathbf{\hat{y}}^+ | \mathcal{\hat{Y}}^-, \mathbf{x}), 
\end{equation}
where $p_{D_\phi}(\mathbf{\hat{y}}^+ | \mathcal{\hat{Y}}^-, \mathbf{x})$ is the probability of selecting $\hat{y}^+$ from $\{\mathbf{\hat{y}}^+\}\cup \mathcal{Y}^-$, which is computed by applying softmax on the ranking scores:
\begin{equation}
    p_{D_\phi}(\mathbf{\hat{y}}^+ | \mathcal{\hat{Y}}^-, \mathbf{x}) = \frac{\text{exp}^{s_{{\hat{y}}^+} }} { \text{exp}^{s_{{\hat{y}}^+} } +  \sum_{\mathbf{\hat{y}}^-\in \hat{\mathcal{Y}}^- } \text{exp}^{s_{\hat{y}^-}}},
\end{equation}
where $s_{{\hat{y}}^+}$ and $s_{\hat{y}^-}$ are the ranking scores of positive candidate and negative candidate, respectively.

After several steps of updating the ranker, we fix the ranker and update the generator.

\textbf{Fix {$D_{\phi}(\cdot)$}, Train $G_{\theta}(\cdot)$}: the generator model is trained in two ways. The first one is $\mathcal{L}_\text{NLL}$, which uses a teacher-forcing mechanism to minimize the negative log-likelihood loss function over the training instances as discussed in Section ~\ref{sec:basic_generator} ( Eqn.~\ref{eq:LNLL}). The second one is $\mathcal{L}_\text{RL}$ - a reinforcement learning-based approach in which the generator model acts as a policy network to produce a list of text samples $\hat{\mathcal{Y}}$ given the input $\mathbf{x}$, and the ranker model gives a reward to each text sample in $\hat{\mathcal{Y}}$ based on its ranking score. The generator model can be trained by maximizing (minimizing) the expected (negative) reward ~\citep{NIPS1999_464d828b}:
\begin{equation}\label{eq:loss_jgr}
    \mathcal{L}_{\text{RL}} = -\sum\limits_{\mathbf{\hat{y}}\in\mathcal{\hat{Y}}} (\mathcal{R}(\mathbf{\hat{y}}) - b) \sum\limits_{t}\log p_{G_{\theta}}(\hat{y}_t|\hat{y}_{<t},\mathbf{x}),
\end{equation}
where $\mathcal{R}(\mathbf{\hat{y}})$ is the reward for sample $\mathbf{\hat{y}}$,  calculated by combining the matching score $\Delta(\mathbf{\hat{y}}, \mathbf{y})$ and the ranking score $s_{\hat{\mathbf{y}}}$: $\mathcal{R}(\mathbf{\hat{y}}) = \Delta(\mathbf{\hat{y}}, \mathbf{y}) + s_{\hat{\mathbf{y}}}$. 
A baseline $b$ is used to reduce the variance in RL training, which is computed by averaging the rewards of all samples in the candidate set:
$   b = \sum\limits_{\mathbf{\hat{y}}\in\mathcal{\hat{Y}}} \mathcal{R}(\mathbf{\hat{y}}) / C$.  We then combine $\mathcal{L}_\text{NLL}$ and $\mathcal{L}_\text{RL}$ to form the final objective function for generator model training :
\begin{equation}
\label{eq:generator_obj}
    \mathcal{L}^{\theta} = \mathcal{L}_\text{NLL} +  \mathcal{L}_{\text{RL}}.
\end{equation}

After updating the generator for several steps, we go back to fixing the generator and updating the ranker. This iteration will continue until the entire JGR framework converges.

\begin{algorithm}[t]
	\caption{Joint Training of Generator and Ranker (JGR)}
	\label{algo:JGR_train}
	\begin{algorithmic}[1]
		\REQUIRE Generator $G_{\theta}$; Ranker $D_{\phi}$; Training data $\mathbb{D}$.
		\STATE Initialize $G_{\theta}$ and $D_{\phi}$ from the pre-trained language models.
		\STATE Train the warm-up generator $G_{\theta}^0$ on $\mathbb{D}$. 
		\WHILE{model has not converged}
		\FOR{training steps $A$}
		\STATE Sample candidates $\mathcal{\hat{Y}} \sim p_{G_{\theta}}( \cdot  | \mathbf{x})$ for each $\mathbf{x}$ in the mini-batch.
		\STATE Select $\mathbf{\hat{y}}^+$ and $\mathcal{\hat{Y}}^-$ from $\mathcal{\hat{Y}}$
		\STATE Update parameters of $D_{\phi}$ with Eq \ref{eq:ranker_obj}.
		\ENDFOR
		\FOR{training steps $B$}
		\STATE Sample candidates $\mathcal{\hat{Y}} \sim p_{G_{\theta}}( \cdot  | \mathbf{x})$ for each $\mathbf{x}$ in the mini-batch.
		\STATE Compute rewards $\mathcal{R}(\mathbf{\hat{y}})$ for each $\mathbf{\hat{y}} \in \mathcal{\hat{Y}}$.
		\STATE Update parameters of $G_{\theta}$ with Eq \ref{eq:generator_obj}.
		\ENDFOR
		\ENDWHILE
		
	\end{algorithmic}  
\end{algorithm}

\begin{table*}[t]
	\centering
	
        \vspace{-0.2cm}
	\resizebox{0.98\textwidth}{!}{
	\begin{tabular}{lcccccccc}
		\toprule
		\multirow{2}*{Method} & \multicolumn{4}{c}{CNN/DailyMail}  &\multicolumn{4}{c}{SAMSum}\\ 
		&R-1& R-2 & R-L&AVG  &R-1& R-2 & R-L & AVG\\
		\hline
		Lead-3 & 40.42& 17.62 &36.67 & \cellcolor[HTML]{ECF4FF}31.57& - &- &- &\cellcolor[HTML]{ECF4FF}-\\
		PTGEN~\citep{see2017get} & 36.44& 15.66 &33.42 & \cellcolor[HTML]{ECF4FF}28.51 & - &- &- &\cellcolor[HTML]{ECF4FF}-\\
		PTGEN-COV~\citep{see2017get} & 39.53& 17.28 &36.38 & \cellcolor[HTML]{ECF4FF}31.06 & - &- &- &\cellcolor[HTML]{ECF4FF}-\\
		BART~\citep{lewis2019bart} & 44.16$^\ast$& 21.28$^\ast$ &40.90$^\ast$ & \cellcolor[HTML]{ECF4FF}35.45$^\ast$  & 52.86$^{\dag\ast}$ & 28.24$^{\dag\ast}$  & 48.57$^{\dag\ast}$  & \cellcolor[HTML]{ECF4FF}43.22$^{\dag\ast}$ \\ 
		PEGASUS~\citep{pmlr-v119-zhang20ae} & 44.17  & 21.47 & 41.11 & \cellcolor[HTML]{ECF4FF}35.58  & 51.99  &27.59  &47.56  &\cellcolor[HTML]{ECF4FF}42.38   \\
		ProphetNet~\citep{qi2021prophetnet} & 44.20  & 21.17 & 41.30 & \cellcolor[HTML]{ECF4FF}35.56 &52.62 &27.77$^\dag$ &48.33 &\cellcolor[HTML]{ECF4FF}42.91 \\
		GSUM~\citep{dou-etal-2021-gsum} & 45.94  & 22.32 & 42.48 & \cellcolor[HTML]{ECF4FF}36.91 & - &- &- &\cellcolor[HTML]{ECF4FF}-\\
		BRIO~\citep{liu-etal-2022-brio} & 47.48$^\ast$  & 23.55$^\ast$ & 44.57$^\ast$ & \cellcolor[HTML]{ECF4FF}38.53$^\ast$& - &- &- &\cellcolor[HTML]{ECF4FF}-\\
	\hline
		
		JGR-G& 46.86  & 23.18 & 43.74 & \cellcolor[HTML]{ECF4FF}37.93 &53.85& 29.22 & 49.93&\cellcolor[HTML]{ECF4FF}44.33 \\
		JGR-R& 47.63  & \textbf{23.59} & 44.50 & \cellcolor[HTML]{ECF4FF}38.57  &\textbf{54.30}& \textbf{29.48}& \textbf{50.51}&\cellcolor[HTML]{ECF4FF}\textbf{44.76}\\
        JGR-G$_\text{init\ w.  BRIO}$& 48.39  & 23.22 & 46.11 & \cellcolor[HTML]{ECF4FF}39.24 &-& - & -&\cellcolor[HTML]{ECF4FF}- \\
        JGR-R$_\text{init\ w.  BRIO}$& \textbf{48.86}  & 23.35 & \textbf{46.56} & \cellcolor[HTML]{ECF4FF}\textbf{39.59} &-& -& -&\cellcolor[HTML]{ECF4FF}-\\
		\bottomrule
	\end{tabular}
	}
 \caption{Overall results on CNN/DailyMail and SAMSum. ``JGR-G'' indicates the generator model in JGR, and ``JGR-R'' is using the ranker of JGR to re-rank the outputs of JGR-G. The results with ``$^\dag$'' means from our implementation. The results with ``$^\ast$'' are the results of backbone models for JGR-G.
}
	\label{tab:res_cnndm_xsum}
\end{table*}

\begin{table}[t]
    \centering
    \vspace{-0.2cm}
    \resizebox{0.95\linewidth}{!}{
	\begin{tabular}{lccc}
		\toprule
		
		&R-L& B-4 &MTR \\
		\hline
	    MASS~\citep{pmlr-v97-song19d}& 50.98 & 23.14 & 25.36 \\
		BART~\citep{lewis2019bart}& 51.46$^\ast$& 23.14$^\ast$ & 26.56$^\ast$ \\ 
		UNILM~\citep{NEURIPS2019_c20bb2d9} &52.04 & 23.75 & 25.61 \\
		ProphetNet~\citep{qi2020prophetnet} & 51.50& 22.50& 26.00\\
		\hline
		JGR-G& 52.79& 24.52& 26.46 \\
		JGR-R& \textbf{53.57}& \textbf{24.73}& \textbf{26.97} \\
		\bottomrule
	\end{tabular}
	}
    \caption{Overall results on SQUAD 1.1.}
    \vspace{-0.5cm}
	\label{tab:res_squad}
\end{table}

\section{Experimental Settings}


\subsection{Datasets}
We evaluate the proposed method on four publicly available benchmarks across four domains: CNN/DailyMail~\citep{hermann2015cnndm} for abstractive summarization, SAMSum~\citep{gliwa2019samsum} for conversational summarization, SQuAD 1.1~\citep{rajpurkar2016squad} for question generation, and PersonaChat~\cite{zhang2018personalizing} for dialogue generation. The details of these benchmarks and the used evaluation metrics are given in Appendix \ref{sec:statistic}.

\subsection{Implemention Details}\label{sec:exp_setting}
We use  BART-large~\citep{lewis2019bart} as the backbone model for the generator. The backbone of the ranker is based on RoBERTa-large~\citep{liu2019roberta}. The generator and ranker models are initialized with the off-the-shelf checkpoints\footnote{RoBERTa:~\url{https://huggingface.co/roberta-large}, BART:~\url{https://huggingface.co/facebook/bart-large}}. 
On CNN/DailyMail , apart from initializing JGR with the language models, we also evaluate JGR that initializes the generator using the previous state-of-the-art model BRIO~\citep{liu-etal-2022-brio}.\footnote{BRIO:~ \url{https://github.com/yixinL7/BRIO}}


During training, the generator model adopts a nucleus sampling approach to generate the candidate set with temperature $= 1.0$ and top($p$) $= 1.0$. In inference, we apply beam search decoding strategy with beam size $ = 16$, and length penalty = $1.0$ for the generator, and we take the output text with the highest beam search score as the final output of the generator. We use the ranker to re-rank the total 16 beam search results and pick the one with the highest ranking order as the final output of the ranker. The details of other hyper-parameters (e.g., learning rate and training epochs, etc) are listed in Appendix \ref{sec:hyper_parameter}. JGR is implemented based on the open-source Huggingface Transformers framework~\citep{wolf-etal-2020-transformers}. We conduct experiments on a single node of 8 NVIDIA A100 GPUs.    

\begin{table}[t]
    \centering
        \vspace{-0.2cm}
    \resizebox{0.95\linewidth}{!}{
	\begin{tabular}{lcccc}
		\toprule
		&B-1& B-2 & D-1& D-2\\
		\hline
		BART~\citep{lewis2019bart}&49.9$^\ast$ &40.0$^\ast$ &1.3$^\ast$ &8.0$^\ast$ \\
		PLATO$_{\text{w/o lantent}}$~\citep{bao-etal-2020-plato} &40.6 &31.5 & \textbf{2.1} & \textbf{12.1} \\
		PLATO~\citep{bao-etal-2020-plato} & 45.8 &35.7 & 1.2 & 6.4\\
		ProphetNet~\citep{qi2020prophetnet}  &46.7&39.0 &1.3 & 7.5 \\
		DialogVED~\citep{chen-etal-2022-dialogved} & 48.2 & 39.9 & 1.5 & 9.4 \\
		\hline
		JGR-G&52.5 & 43.2& 1.4 & 6.2 \\
		JGR-R&\textbf{53.3} & \textbf{43.5} & 1.5 & 8.0 \\
		\bottomrule
	\end{tabular}
	}
    \caption{Overall results on PersonaChat.}
    \vspace{-0.5cm}
	\label{tab:res_personachat}
\end{table}
It is worth noting that in order to initialize the ranker with a more general and reasonable ranking function, we increase the number of training steps and add a certain number of warm-up steps at the first ranker training iteration. The hyper-parameters of the first ranker training iteration are also introduced in Appendix~\ref{sec:hyper_parameter}.

\section{Results and Analyses}

\subsection{Overall Results}
Table~\ref{tab:res_cnndm_xsum} shows the results of JGR and other baseline methods on summarization tasks CNN/DailyMail and SAMSum. ``Lead-3'' is an ad-hoc summarization approach that uses the first three sentences in the article as the summary. ``PTGEN'' and ``PTGEN-COV'' are sequence-to-sequence generation methods without pre-training. Other baselines are pre-trained language models fine-tuned on the benchmarks. ``JGR-G'' indicates the generator model in JGR, and ``JGR-R'' is using the ranker of JGR to re-rank the outputs of JGR-G. 
``JGR-G/R$_\text{init\ w.  BRIO}$'' are our JGR with the generator initialized from BRIO.
As shown in Table ~\ref{tab:res_cnndm_xsum}, the generator model (JGR-G) itself achieves a considerable performance gain compared with its backbone models on both the two benchmarks, which verifies the effectiveness of the proposed JGR training to obtain a better generator. On both CNN/DailyMail and SAMSum, the ranker (JGR-R) can further improve the performance of JGR-G. Both JGR-G and JGR-R can reach state-of-the-art on SAMSum. If initialized with BRIO, both our JGR-G and JGR-R can surpass the state-of-the-art on CNN/DailyMail with a considerable margin.

In Table~\ref{tab:res_squad}, we compare the performance of JGR with four pre-trained language models~\citep{pmlr-v97-song19d,lewis2019bart,NEURIPS2019_c20bb2d9,qi2020prophetnet} on SQuAD 1.1, since they have reported the results finetuned and evaluated in the same data split as in \citet{liu2020glge}. 
With a relatively weak backbone model, BART, our JGR-G can still outperform all the compared baselines. And JGR-R can also further improve the results of JGR-G.

Table \ref{tab:res_personachat} shows the results of compared methods in persona-based response generation. 
As shown in the results, our JGR-G and JGR-R can surpass the baselines significantly on the metrics of BLEU-1 and BLEU-2. However, both JGR-G and JGR-R can only perform the same level of the baselines on Distinct-1 and Distinct-2. 
It is noteworthy that PLATO and DialogVED are the only two language models that are pre-trained using a conversational corpus among these baselines. They achieved high scores on Distinct-1 and Distinct-2, showing the importance of pre-training corpus.

\begin{table}[t]
    \centering
	\resizebox{0.9\linewidth}{!}{
	\begin{tabular}{l|l|cccc}
		\toprule
		Generator &Ranker &R-1& R-2 &R-L &Gain\\
		\hline
		$G^0$&- &44.16& 21.28 &40.90 &0.00\\
		$G^0$& SimCLS&46.67 &22.15 &43.54 &2.00\\
		$G^0$& RefSum&45.15 &21.70 &42.00 &0.83\\ 
		$G^0$& SumRanker&46.62 &22.39 &43.59 &2.08\\
		$G^0$& BRIO&47.28 &22.93 &44.15 &\textbf{3.84}\\
		$G^0$& COLO&46.33 &22.15 &43.08 &1.73\\
		$G^0$& $D^0$&45.54 &22.27 &42.25 &1.24\\ 
		\hline
		JGR-G& -& 46.86  & 23.18 & 43.74 & 0.00 \\
		JGR-G& JGR-R&\textbf{47.63}  & \textbf{23.59} & \textbf{44.50} & 0.64\\ 
		\bottomrule
	\end{tabular}
	}
	\caption{The results of different generate-then-rank frameworks on CNN/DailyMail. ``Gain'' represents the performance gain of ranker compared with the used generator over the average score.}
	\label{tab:res_ranker}
	\vspace{-0.3cm}
\end{table}

\subsection{Performance of Generate-then-Rank Frameworks}\label{app:ranker}
Recently, several works adopt the generate-then-rank framework, especially on the summarization tasks~\citep{liu-liu-2021-simcls,liu-etal-2021-refsum,ravaut2022summareranker,liu-etal-2022-brio,an-etal-2022-colo}. Different from JGR, these methods do not introduce the iterative training of the generator and ranker. We compare these methods with that our JGR-R on CNN/DailyMail. Since all the above methods train the ranker separately with the fine-tuned BART as the generator on CNN/DailyMail, we only report their results in this setting.

The experimental results are shown in Table \ref{tab:res_ranker}, where $G^0$ denotes the base generator, i.e. BART, and $D^0$ is the ranker after the first ranker training iteration, as described in Section \ref{sec:exp_setting}. Several observations can be seen in the results. First, our JGR achieves the highest score with the inference pipeline. Second, on CNN/DailyMail, the performance gain brought by JGR-R is not as big as other related methods which introduced some extra modules to their models. 
Third, on CNN/DailyMail, after the joint training in JGR, the performance gain brought by the ranker drops. We think this is because as the generator's performance grows, the quality of candidates rises, causing the ranker harder to pick the best among all candidates.

\subsection{Impact of Rewards}\label{sec:res_reward}

In this section, we investigate the impact of rewards. We compare different reward settings on CNN/DailyMail. The compared methods are as follows: 1) \textbf{Self-critic} is the conventional self-critical reinforcement-learning method where the rewards are the metric scores $\Delta(\mathbf{\hat{y}}, \mathbf{y})$, and the greedy search output is used as baseline~\cite{8099614, paulus2018a}. 2) \textbf{Actor-critic} is the RL-based method that trains a critical model to fit the metric scores $\Delta(\mathbf{\hat{y}}, \mathbf{y})$, and uses the critical score as the reward to train generator~\citep{konda1999actor, bahdanau2017an, le2022coderl}.
3) \textbf{JGR-G}$_{\text{only mr}}$/\textbf{JGR-G}$_{\text{only rr}}$ are our JGR where the generator is trained without the rewards from generator/metrics. The standard NLL loss is added in all the compared methods. The results are shown in Table \ref{tab:res_reward}.

\begin{table}[t]
    \centering
    \resizebox{0.9\linewidth}{!}{
	\begin{tabular}{lcccccc}
		\toprule
		&R-1& R-2 & R-L& AVG\\
		\hline
		BART & 44.16& 21.28 &40.90 & \cellcolor[HTML]{ECF4FF}35.45\\
		\hline
		Self-critic&44.14 &21.20 &40.95 &\cellcolor[HTML]{ECF4FF}35.43 \\
            Actor-critic&45.04 &21.99 &41.71 &\cellcolor[HTML]{ECF4FF}36.25 \\
		\hline
		JGR-G&\textbf{46.86} &\textbf{23.18} &43.74 &\cellcolor[HTML]{ECF4FF}\textbf{37.93} \\
		JGR-G$_{\text{only mr}}$&44.20 &21.37 &41.04 &\cellcolor[HTML]{ECF4FF}35.54 \\
		JGR-G$_{\text{only rr}}$&46.76 &22.99 &\textbf{43.81} &\cellcolor[HTML]{ECF4FF}37.85 \\
		\bottomrule
	\end{tabular}
	}
	\captionof{table}{Results generator trained with different type of rewards on CNN/DailyMail.}
        \vspace{-0.4cm}
	\label{tab:res_reward}
\end{table}

\begin{figure}[t]
\centering
    \includegraphics[width = 0.92\linewidth]{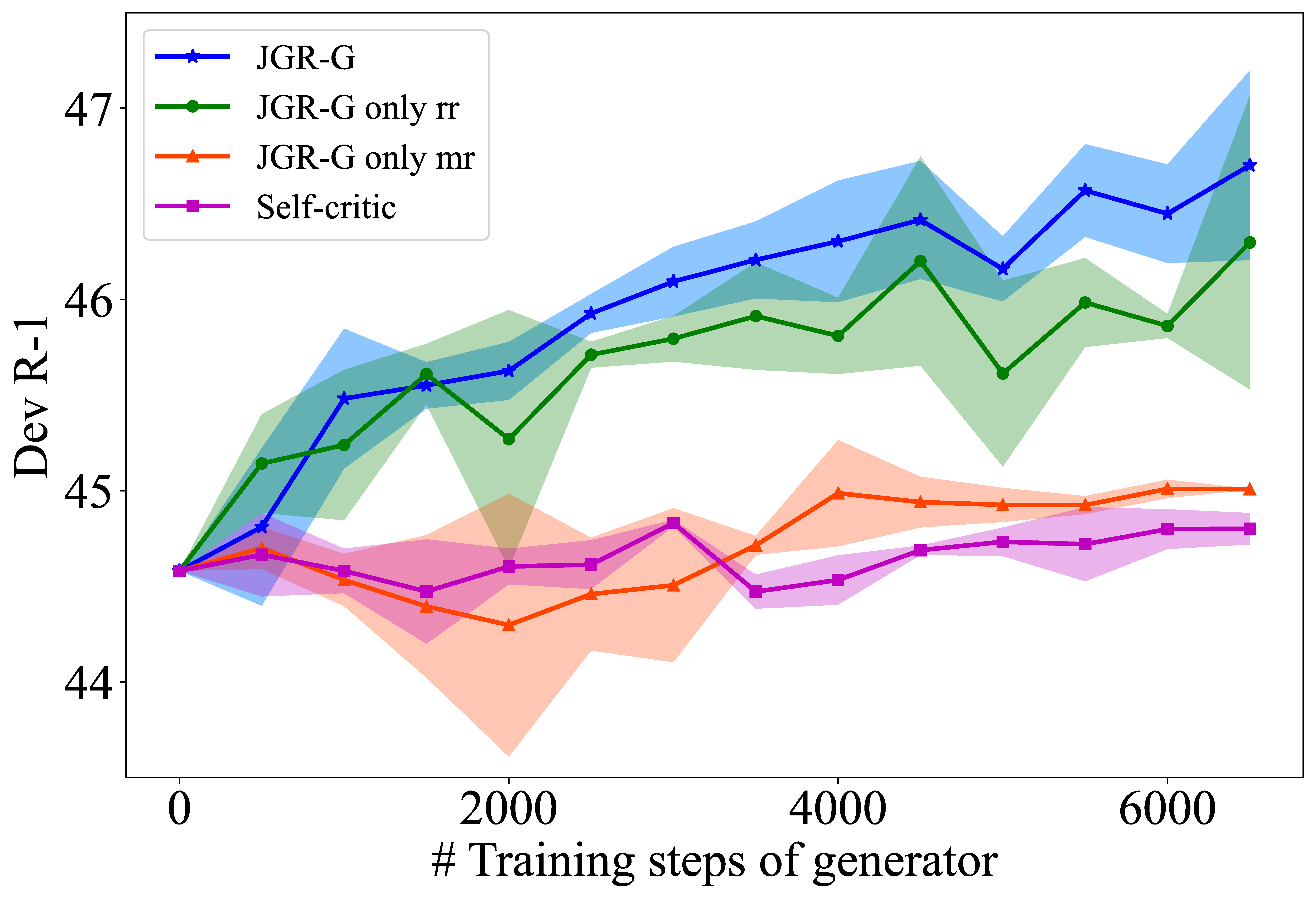}
    \vspace{-0.3cm}
    \caption{Dev scores on CNN/DailyMail with three random runs for methods with different types of rewards. }
    \vspace{-0.5cm}
\label{fig:dev-scores}
\end{figure}

According to the results, our JGR can outperform traditional RL significantly. 
Both JGR-G$_{\text{only mr}}$ and JGR-G$_{\text{only rr}}$ suffer a performance decline compared to standard JGR-G, and the performance of JGR-G$_{\text{only mr}}$ is far worse than that of JGR-G$_{\text{only rr}}$. In addition, the Actor-critical method outperforms the Self-critical method. The above two observations indicate that using rewards from a trained reward model contributes more than using rewards from metrics, and it is better to combine them. In Figure \ref{fig:dev-scores}, we plot the curves of the dev scores under 3 random runs for the compared methods. As illustrated in the figure, although the standard Self-critical method appears to have a small variance under different random runs, its dev scores are hard to grow while training. The JGR-G$_{\text{only rr}}$ has a smaller variance than JGR-G$_{\text{only mr}}$, however, it fails to achieve a high dev score. Our standard JGR, which combines metric rewards and ranker rewards, not only shows the relatively small variance in randomized trials but also can steadily improve the dev score during training.

\subsection{Candidate Picking Strategies}
We examine how different types and numbers of candidates can affect the performance of JGR. We first compare different methods of picking positive candidates and negative candidates when training the ranker. The results are shown in Table~\ref{tab:res_cand_pick}. The $\mathbf{\hat{y}}^+$=GT denotes the positive candidate $\mathbf{\hat{y}}^+$  being always the reference, not the generated samples. The result shows that if the best candidate is always the reference, the performance of the generator is not as good as the standard JGR, and the ranker's performance is even worse than the generator. This is because the ranker is misled by the reference, thus it may always misclassify the references as the positive candidates, while other candidates sampled by the generator as the negative candidates. As a result, neither the ranker is well-trained, nor it can pass proper rewards to train the generator. 
\begin{table}[t]
	\centering
	\resizebox{0.98\linewidth}{!}{
	\begin{tabular}{lcccccccc}
		\toprule
		\multirow{2}*{ }&\multicolumn{4}{c}{Generator}&\multicolumn{4}{c}{Ranker}\\
		&R-1& R-2 & R-L& AVG &R-1& R-2 & R-L &AVG\\
		\hline
		$\mathbf{\hat{y}}^+$=GT & 45.64& 22.27& 42.55& 36.82  & 44.20& 21.46& 41.22& 35.63\\
		\hline
		$\hat{\mathcal{Y}}^-=\text{BOT}(\hat{\mathcal{Y}})$&\textbf{46.86}& \textbf{23.18} & \textbf{43.74} &\textbf{37.93} &\textbf{47.63} &\textbf{23.59} &\textbf{44.50} &\textbf{38.57} \\
		$\hat{\mathcal{Y}}^-=\text{TOP}(\hat{\mathcal{Y}})$&44.16 &21.31 &41.00 &35.49 &44.07 &21.23 &40.91 & 35.40\\
		$\hat{\mathcal{Y}}^-=\text{RAND}(\hat{\mathcal{Y}})$&44.68 &21.65 &41.42 &35.92 & 45.80&22.68 &42.56 &37.01 \\
		$\hat{\mathcal{Y}}^-=\text{TOP-BOT}(\hat{\mathcal{Y}})$&44.86 & 21.80& 41.64&36.10 &46.12 & 22.76& 42.91& 37.26\\
		\bottomrule
	\end{tabular}
	}
	\caption{Results of JGR with different candidate picking strategies on CNN/DailyMail. }
	\label{tab:res_cand_pick}
	\vspace{-0.3cm}
\end{table}

\begin{table}[t]
	\centering
	\resizebox{0.95\linewidth}{!}{
	\begin{tabular}{lcccccccc}
		\toprule
		\multirow{2}*{ }&\multicolumn{4}{c}{Generator}&\multicolumn{4}{c}{Ranker}\\
		&R-1& R-2 & R-L& AVG &R-1& R-2 & R-L &AVG\\
		\hline
		$C=2$& 44.59& 21.63& 41.32& 35.85& 46.15& 22.76& 42.86& 37.26\\
		$C=4$& 45.44& 22.19& 42.80& 36.81& 46.70& 23.10& 43.81& 37.87\\
		$C=6$& 46.36& 22.77& 42.94& 37.37& 47.32& 23.49& 44.29& 38.37\\
		$C=8$ & \textbf{46.86}  & \textbf{23.18} & \textbf{43.74} & \textbf{37.93}&\textbf{47.63}  & 23.59 & \textbf{44.50}& \textbf{38.57}\\
		$C=16$&  46.34& 22.97 &43.11 & 37.47  &47.34  &\textbf{23.64} &44.13  &38.37  \\
		$C=32$&  46.14& 22.78 &42.87 & 37.26  &47.25  &23.48 &43.98  &38.24  \\
		$C=40$& 46.29& 22.98 & 43.00 & 37.42 & 47.26 & 23.60& 44.00 &38.29 \\
		\bottomrule
	\end{tabular}
	}
	\caption{Results of JGR with numbers of sampled candidates on CNN/DailyMail. }
	\label{tab:res_cand_num}
	\vspace{-0.5cm}
\end{table}
The last four lines of Table~\ref{tab:res_cand_pick} show the results of methods for picking negative samples, i.e., with the lowest matching scores ($\text{BOT}(\hat{\mathcal{Y}})$, our standard setting), with the highest matching scores ($\text{TOP}(\hat{\mathcal{Y}})$), randomly pick ($\text{RAND}(\hat{\mathcal{Y}})$), and half has the highest matching scores and the second half has the lowest matching score ($\text{TOP-BOT}(\hat{\mathcal{Y}})$). From the results, we can see that our standard setting ($\text{BOT}(\hat{\mathcal{Y}})$) significantly outperforms other negative candidate picking strategies.

In Table~\ref{tab:res_cand_num}, we show the performance of JGR with different numbers of sampled candidates when training the generator. According to the results, under a certain range ($C=2\sim8$), the performance of JGR goes up as the number of candidates increases. We attribute this to the fact that increasing the number of candidates means that the generator can be optimized on more probabilities from candidates, which is to some extent a way of data augmentation. However, the performance does not grow as desired when the number of candidates becomes too large.


\subsection{Advanced Metrics and Human Evaluation}\label{sec:advanced_metric}

A model trained with RL objective may succeed in the metrics it uses as the reward function but perform poorly in other metrics. We hope to investigate whether JGR, which uses the RL objective to train its generator, suffers from the same problem. Firstly, we use three advanced metrics, namely BERTScore~\citep{Zhang*2020BERTScore:}, FactCC~\citep{kryscinski-etal-2020-evaluating}, and QuestEval~\citep{scialom-etal-2021-questeval}, to evaluate JGR on CNN/DailyMail. BERTScore measures the semantic similarity of the predicted summary and ground-truth reference. FactCC and QuestEval use a trained language model to measure the factual consistency between the generated summary and input source document. According to the results shown in Table \ref{tab:res_advanced}, JGR-G and JGR-R both achieve higher BERTScore than BART, indicating that they can generate summaries with better semantic quality. For FactCC and QuestEval, which measure factual consistency, JGR-G and JGR-R also surpass the BART baseline. 

\begin{table}[t]
	\centering
    \resizebox{0.9\linewidth}{!}{
	\begin{tabular}{lcccccc}
		\toprule
		&BERTScore& FactCC & QE\\
		\hline
		BART & 88.47 & 57.54 &50.56 \\
		\hline
		JGR-G&88.90 & 60.33   &52.09  \\
		JGR-R&\textbf{88.96} & \textbf{61.59} &\textbf{52.18}  \\

		\bottomrule
	\end{tabular}
	}
	\captionof{table}{Performance on BERTScore, FactCC, and QuestEval.}
	\label{tab:res_advanced}
	
	\vspace{-0.3cm}
\end{table}

\begin{table}[t]
	\centering
    \resizebox{0.85\linewidth}{!}{
	\begin{tabular}{lcccccc}
		\toprule
		&JGR-G wins& Tie & JGR-G loses\\
		\hline
		Inform. & 58 & 3 &39 \\
		Fact.&61 & 7   &32 \\
		Read&45 & 15 &40  \\

		\bottomrule
	\end{tabular}
	}
	\captionof{table}{Result of human evaluation compared with BART.}
	\label{tab:res_human}
	
	\vspace{-0.5cm}
\end{table}



We also conduct a human evaluation on CNN/DailyMail\footnote{More details about human evaluation are in Appendix~\ref{app:human-eval}.}. Following Blenderbot v2~\citep{roller-etal-2021-recipes}, we randomly picked 100 cases from the CNN/DailyMail test set and asked the annotators to explicitly compare which generated text is better for each pair of summaries generated by JGR-G and BART, rather than assign an evaluation score. This explicit comparison can avoid the per annotator bias in numerical scores (e.g., annotators who tend to give generous scores), and remedy many of the issues of sequential effects such as contrasting with a previous case. Three aspects corresponding to the generation quality are evaluated, namely informativeness (Inform.), factual consistency (Fact.), and readability (Read.). As shown in Table~\ref{tab:res_human}, JGR-G beats BART in 58 cases w.r.t informativeness and 61 cases w.r.t. factual consistency, indicating that JGR-G performs better than BART on informativeness and factual consistency. For readability, JGR can generate summaries as readable as BART. 

To conclude, though trained with reinforcement learning aimed at optimizing ROUGE score, JGR still does not compromise on other aspects of summary quality, including semantic similarity, factual consistency, informativeness, and readability.

\begin{table}[t]
	\centering
	\resizebox{0.95\linewidth}{!}{
	\begin{tabular}{lcccccccc}
		\toprule
		\multirow{2}*{ }&\multicolumn{4}{c}{Generator}&\multicolumn{4}{c}{Ranker}\\
		&R-1& R-2 & R-L& AVG &R-1& R-2 & R-L& AVG\\
		\hline
		JGR& \textbf{46.86 }  & \textbf{23.18} & \textbf{43.74} & \textbf{37.93} &\textbf{47.63}  & \textbf{23.59} & \textbf{44.50}& \textbf{38.57}\\
		\ \ w/o joint& 45.02& 21.83& 42.40&36.42 & 45.10& 21.81& 42.47 & 36.46\\ 
		\bottomrule
	\end{tabular}
	}
	\caption{Results of JGR and JGR without joint training on CNN/DailyMail. }
	\label{tab:joint}
	
	\vspace{-0.3cm}
\end{table}

\begin{figure}[t] \centering    
\subfigure[Wasserstein distances] { 
\label{fig:wasserstein}     
\includegraphics[width=0.46\linewidth]{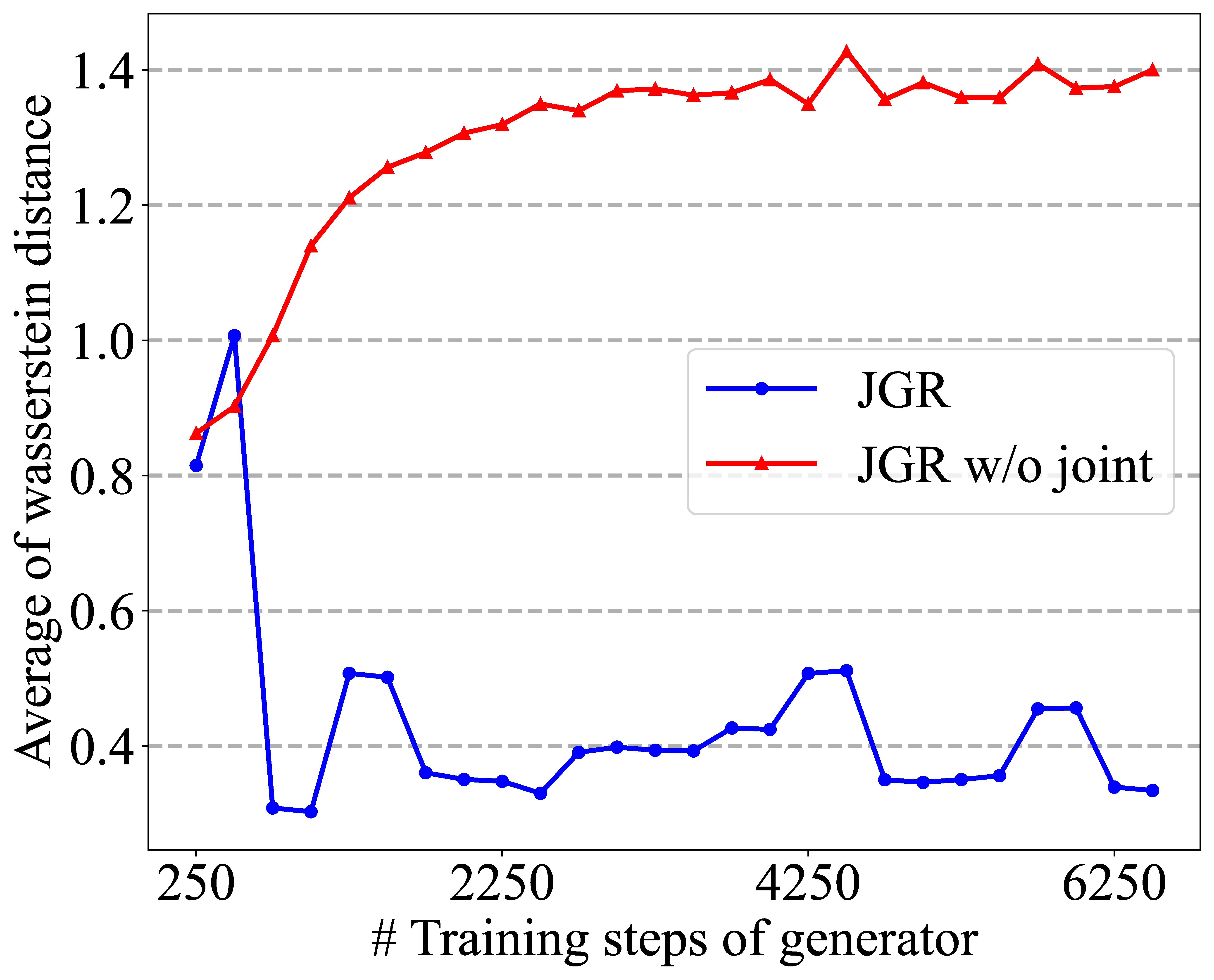}     
}   
\subfigure[Self-BLEU] {
 \label{fig:std-rewards}     
\includegraphics[width=0.46\linewidth]{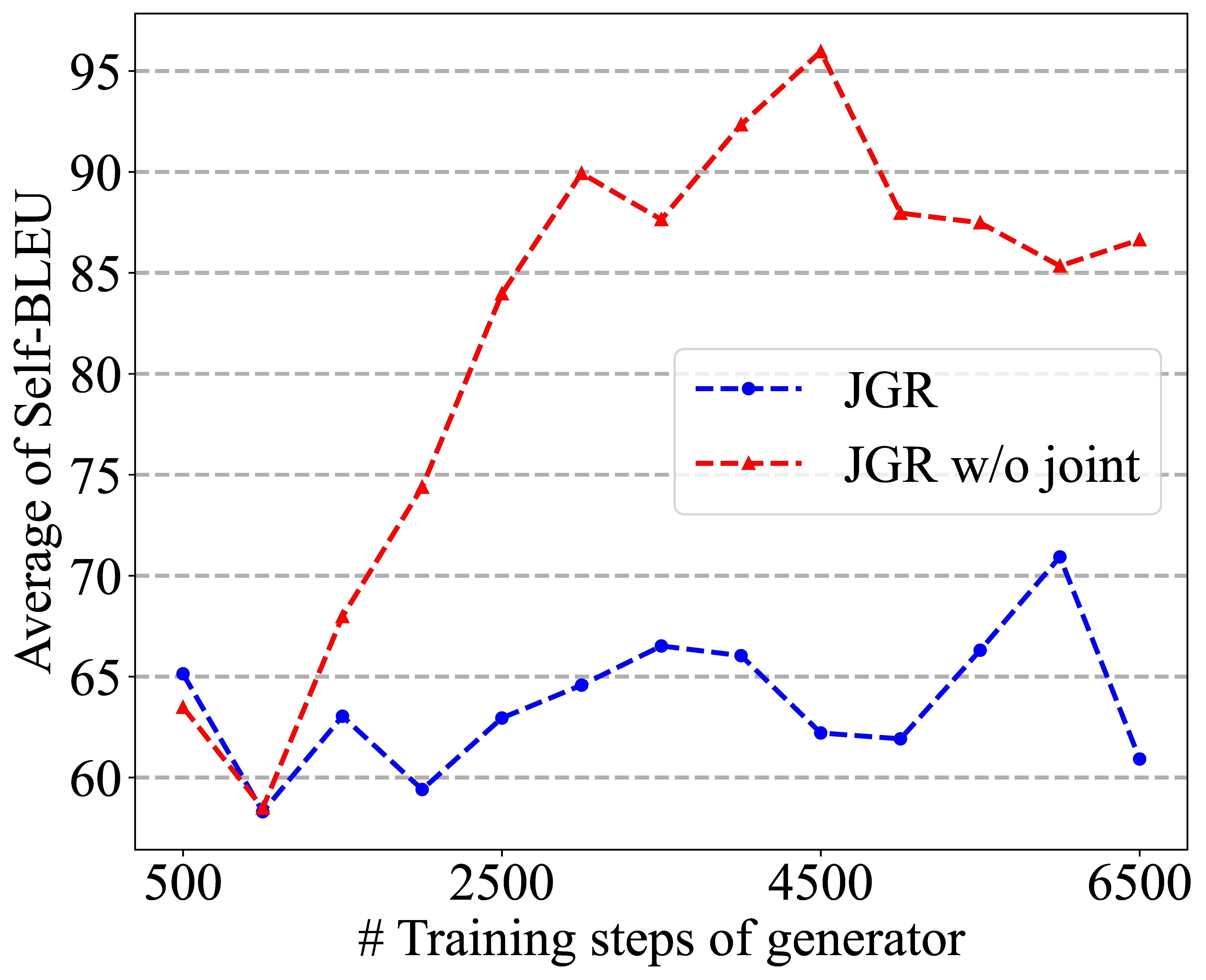}  
}

\caption{(a) The average of Wasserstein distances between ranker rewards and metrics rewards. (b) The average of self-BLEU at each training interval.}     
\vspace{-0.5cm}
\end{figure}

\subsection{Does Joint Training Matter?}\label{sec:joint_train}

To see how our proposed joint (iterative) training of the generator and ranker affects JGR, we compare the performance of our JGR and the variant that trains the generator in the same reinforcement learning paradigm as the JGR while fixing the ranker after fully training it (JGR$_{\text{w/o joint}}$)\footnote{More details are given in Appendix~\ref{app:without-joint}.}. As the results shown in Table \ref{tab:joint}, JGR$_{\text{w/o joint}}$ is far worse than JGR, and JGR-R$_{\text{w/o joint}}$ achieves no performance gain over JGR-G$_{\text{w/o joint}}$, which indicates the importance of the iterative training. 
To take an in-depth look, we analyze the distribution of rewards. We first draw the curves of the Wasserstein distance between ranker rewards and metrics rewards at each training interval for JGR and JGR$_{\text{w/o joint}}$. As illustrated in Figure \ref{fig:wasserstein}, the Wasserstein distances of JGR are hovering within a range, while the Wasserstein distances of JGR$_{\text{w/o joint}}$ are growing extremely high, which means the distribution of ranker rewards and metrics rewards are quite different in JGR$_{\text{w/o joint}}$. Therefore we think that JGR-R$_{\text{w/o joint}}$ might not assign the proper rewards to the sampled candidates, due to it not being jointly trained.

We also analyze the diversity of sampled candidates for JGR-G and JGR-G$_{\text{w/o joint}}$. We use self-BLEU\footnote{We introduce the computation of self-BLEU in Appendix~\ref{app:self_bleu}.} to measure the diversity of sampled candidates.
A larger self-BLEU score means a lower diversity of the sampled candidates. We show the curves of the average self-BLEU score for generated candidates at each training interval in Figure \ref{fig:std-rewards}. From the figure, we can see that the self-BLEU of JGR$_{\text{w/o joint}}$ increases rapidly after the generator is trained 1000 steps, while the same situation never happens in JGR. It indicates that if the ranker is not jointly trained with the generator, the rewards it feeds back to the generator will cause the generator to sample candidates that are more and more similar to each other, making the training of JGR harder. On the contrary, joint training can erase this phenomenon and help to keep a certain level of diversity in sampled candidates, thus leading to better training.

\subsection{More Discussions}

Due to the page limit, we show more discussions about JGR compared to reinforcement learning, GAN, data augmentation in Appendix~\ref{app:discussion}, the impact of decoding strategies in Appendix~\ref{app:decode}.

\section{Conclusion}

In this paper, we propose a novel Joint training of Generator and Ranker framework, namely JGR, for natural language generation. Both the generator and ranker of our JGR can achieve state-of-the-art results on several benchmarks in the areas of summarization, question generation, and dialog. We also analyze our JGR in several aspects and find that: First, the rewards from the ranker work better than the rewards from the direct metrics such as BLEU, but combining them together helps the training become more stable. Second, during training, letting the ranker be trained on the candidates generated by the generator exclusively is even better than previous approaches using ground-truth as positive examples. Third, more candidates being sampled during training can lead to better performance, which is consistent with the findings from data augmentation. Fourth, though trained with reinforcement learning aimed at optimizing automatic evaluation metrics, JGR still does not compromise on other aspects of generation quality. Finally, the joint training paradigm helps the generator sample candidates with higher diversity, which in turn contribute to the training. 


\section*{Limitations}
So far JGR has only been evaluated on the domains of summarization, conversational summarization, question generation, and dialog. It should be evaluated on a wider range of benchmarks, such as machine translation and code generation. And we have not explored JGR's performance with extra-large language models such as GPT-3. We will evaluate JGR on the above points in the future.

Because the generator of JGR samples candidates using auto regressive sampling, it may occupy relatively longer computational time and larger memory then the conventional MLE training. Though the performance of JGR is satisfactory, we still want to improve its computational costs. We will try non-auto regressive sampling and other improvements such as parameter sharing in the future.

\section*{Ethics Statement}
All the experiments are conducted on publicly available datasets, which don't include any private information. Our work doesn't involve identity characteristics or any gender and racial discrimination. 

\bibliography{custom}

\begin{thebibliography}{56}
\expandafter\ifx\csname natexlab\endcsname\relax\def\natexlab#1{#1}\fi

\bibitem[{An et~al.(2022)An, Zhong, Wu, Zhu, Huang, and
  Qiu}]{an-etal-2022-colo}
Chenxin An, Ming Zhong, Zhiyong Wu, Qin Zhu, Xuanjing Huang, and Xipeng Qiu.
  2022.
\newblock \href {https://aclanthology.org/2022.coling-1.508} {{C}o{L}o: A
  contrastive learning based re-ranking framework for one-stage summarization}.
\newblock In \emph{Proceedings of the 29th International Conference on
  Computational Linguistics}, pages 5783--5793, Gyeongju, Republic of Korea.
  International Committee on Computational Linguistics.

\bibitem[{Bahdanau et~al.(2017)Bahdanau, Brakel, Xu, Goyal, Lowe, Pineau,
  Courville, and Bengio}]{bahdanau2017an}
Dzmitry Bahdanau, Philemon Brakel, Kelvin Xu, Anirudh Goyal, Ryan Lowe, Joelle
  Pineau, Aaron Courville, and Yoshua Bengio. 2017.
\newblock \href {https://openreview.net/forum?id=SJDaqqveg} {An actor-critic
  algorithm for sequence prediction}.
\newblock In \emph{International Conference on Learning Representations}.

\bibitem[{Bao et~al.(2020)Bao, He, Wang, Wu, and Wang}]{bao-etal-2020-plato}
Siqi Bao, Huang He, Fan Wang, Hua Wu, and Haifeng Wang. 2020.
\newblock \href {https://doi.org/10.18653/v1/2020.acl-main.9} {{PLATO}:
  Pre-trained dialogue generation model with discrete latent variable}.
\newblock In \emph{Proceedings of the 58th Annual Meeting of the Association
  for Computational Linguistics}, pages 85--96, Online. Association for
  Computational Linguistics.

\bibitem[{Chen et~al.(2022)Chen, Gong, Wang, Yao, Qi, Wei, Hu, Zhou, Mao, Chen,
  Cheng, and Duan}]{chen-etal-2022-dialogved}
Wei Chen, Yeyun Gong, Song Wang, Bolun Yao, Weizhen Qi, Zhongyu Wei, Xiaowu Hu,
  Bartuer Zhou, Yi~Mao, Weizhu Chen, Biao Cheng, and Nan Duan. 2022.
\newblock \href {https://doi.org/10.18653/v1/2022.acl-long.333} {{D}ialog{VED}:
  A pre-trained latent variable encoder-decoder model for dialog response
  generation}.
\newblock In \emph{Proceedings of the 60th Annual Meeting of the Association
  for Computational Linguistics (Volume 1: Long Papers)}, pages 4852--4864,
  Dublin, Ireland. Association for Computational Linguistics.

\bibitem[{Cobbe et~al.(2021)Cobbe, Kosaraju, Bavarian, Chen, Jun, Kaiser,
  Plappert, Tworek, Hilton, Nakano, Hesse, and
  Schulman}]{https://doi.org/10.48550/arxiv.2110.14168}
Karl Cobbe, Vineet Kosaraju, Mohammad Bavarian, Mark Chen, Heewoo Jun, Lukasz
  Kaiser, Matthias Plappert, Jerry Tworek, Jacob Hilton, Reiichiro Nakano,
  Christopher Hesse, and John Schulman. 2021.
\newblock \href {https://doi.org/10.48550/ARXIV.2110.14168} {Training verifiers
  to solve math word problems}.

\bibitem[{Cohen and Beck(2019)}]{pmlr-v97-cohen19a}
Eldan Cohen and Christopher Beck. 2019.
\newblock \href {https://proceedings.mlr.press/v97/cohen19a.html} {Empirical
  analysis of beam search performance degradation in neural sequence models}.
\newblock In \emph{Proceedings of the 36th International Conference on Machine
  Learning}, volume~97 of \emph{Proceedings of Machine Learning Research},
  pages 1290--1299. PMLR.

\bibitem[{Dong et~al.(2019)Dong, Yang, Wang, Wei, Liu, Wang, Gao, Zhou, and
  Hon}]{NEURIPS2019_c20bb2d9}
Li~Dong, Nan Yang, Wenhui Wang, Furu Wei, Xiaodong Liu, Yu~Wang, Jianfeng Gao,
  Ming Zhou, and Hsiao-Wuen Hon. 2019.
\newblock \href
  {https://proceedings.neurips.cc/paper/2019/file/c20bb2d9a50d5ac1f713f8b34d9aac5a-Paper.pdf}
  {Unified language model pre-training for natural language understanding and
  generation}.
\newblock In \emph{Advances in Neural Information Processing Systems},
  volume~32. Curran Associates, Inc.

\bibitem[{Dou et~al.(2021)Dou, Liu, Hayashi, Jiang, and
  Neubig}]{dou-etal-2021-gsum}
Zi-Yi Dou, Pengfei Liu, Hiroaki Hayashi, Zhengbao Jiang, and Graham Neubig.
  2021.
\newblock \href {https://doi.org/10.18653/v1/2021.naacl-main.384} {{GS}um: A
  general framework for guided neural abstractive summarization}.
\newblock In \emph{Proceedings of the 2021 Conference of the North American
  Chapter of the Association for Computational Linguistics: Human Language
  Technologies}, pages 4830--4842, Online. Association for Computational
  Linguistics.

\bibitem[{Du et~al.(2017)Du, Shao, and Cardie}]{du2017learning}
Xinya Du, Junru Shao, and Claire Cardie. 2017.
\newblock \href {https://doi.org/10.18653/v1/P17-1123} {Learning to ask: Neural
  question generation for reading comprehension}.
\newblock In \emph{Proceedings of the 55th Annual Meeting of the Association
  for Computational Linguistics (Volume 1: Long Papers)}, pages 1342--1352,
  Vancouver, Canada. Association for Computational Linguistics.

\bibitem[{Gehring et~al.(2016)Gehring, Auli, Grangier, and
  Dauphin}]{gehring2016convolutional}
Jonas Gehring, Michael Auli, David Grangier, and Yann~N Dauphin. 2016.
\newblock A convolutional encoder model for neural machine translation.
\newblock \emph{arXiv preprint arXiv:1611.02344}.

\bibitem[{Gliwa et~al.(2019)Gliwa, Mochol, Biesek, and Wawer}]{gliwa2019samsum}
Bogdan Gliwa, Iwona Mochol, Maciej Biesek, and Aleksander Wawer. 2019.
\newblock Samsum corpus: A human-annotated dialogue dataset for abstractive
  summarization.
\newblock In \emph{Proceedings of the 2nd Workshop on New Frontiers in
  Summarization}, pages 70--79.

\bibitem[{Goodfellow et~al.(2014)Goodfellow, Pouget-Abadie, Mirza, Xu,
  Warde-Farley, Ozair, Courville, and Bengio}]{NIPS2014_5ca3e9b1}
Ian Goodfellow, Jean Pouget-Abadie, Mehdi Mirza, Bing Xu, David Warde-Farley,
  Sherjil Ozair, Aaron Courville, and Yoshua Bengio. 2014.
\newblock \href
  {https://proceedings.neurips.cc/paper/2014/file/5ca3e9b122f61f8f06494c97b1afccf3-Paper.pdf}
  {Generative adversarial nets}.
\newblock In \emph{Advances in Neural Information Processing Systems},
  volume~27. Curran Associates, Inc.

\bibitem[{Hermann et~al.(2015)Hermann, Kocisky, Grefenstette, Espeholt, Kay,
  Suleyman, and Blunsom}]{hermann2015cnndm}
Karl~Moritz Hermann, Tomas Kocisky, Edward Grefenstette, Lasse Espeholt, Will
  Kay, Mustafa Suleyman, and Phil Blunsom. 2015.
\newblock \href
  {https://proceedings.neurips.cc/paper/2015/hash/afdec7005cc9f14302cd0474fd0f3c96-Abstract.html}
  {Teaching machines to read and comprehend}.
\newblock In \emph{Advances in Neural Information Processing Systems 28: Annual
  Conference on Neural Information Processing Systems 2015, December 7-12,
  2015, Montreal, Quebec, Canada}, pages 1693--1701.

\bibitem[{Karpukhin et~al.(2020)Karpukhin, O{\u{g}}uz, Min, Lewis, Wu, Edunov,
  Chen, and Yih}]{karpukhin2020dense}
Vladimir Karpukhin, Barlas O{\u{g}}uz, Sewon Min, Patrick Lewis, Ledell Wu,
  Sergey Edunov, Danqi Chen, and Wen-tau Yih. 2020.
\newblock Dense passage retrieval for open-domain question answering.
\newblock \emph{arXiv preprint arXiv:2004.04906}.

\bibitem[{Konda and Tsitsiklis(1999)}]{konda1999actor}
Vijay Konda and John Tsitsiklis. 1999.
\newblock Actor-critic algorithms.
\newblock \emph{Advances in neural information processing systems}, 12.

\bibitem[{Kryscinski et~al.(2020)Kryscinski, McCann, Xiong, and
  Socher}]{kryscinski-etal-2020-evaluating}
Wojciech Kryscinski, Bryan McCann, Caiming Xiong, and Richard Socher. 2020.
\newblock \href {https://doi.org/10.18653/v1/2020.emnlp-main.750} {Evaluating
  the factual consistency of abstractive text summarization}.
\newblock In \emph{Proceedings of the 2020 Conference on Empirical Methods in
  Natural Language Processing (EMNLP)}, pages 9332--9346, Online. Association
  for Computational Linguistics.

\bibitem[{Lamprier et~al.(2022)Lamprier, Scialom, Chaffin, Claveau, Kijak,
  Staiano, and Piwowarski}]{lamprier2022generative}
Sylvain Lamprier, Thomas Scialom, Antoine Chaffin, Vincent Claveau, Ewa Kijak,
  Jacopo Staiano, and Benjamin Piwowarski. 2022.
\newblock Generative cooperative networks for natural language generation.
\newblock \emph{arXiv preprint arXiv:2201.12320}.

\bibitem[{Le et~al.(2022)Le, Wang, Gotmare, Savarese, and Hoi}]{le2022coderl}
Hung Le, Yue Wang, Akhilesh~Deepak Gotmare, Silvio Savarese, and Steven~CH Hoi.
  2022.
\newblock Coderl: Mastering code generation through pretrained models and deep
  reinforcement learning.
\newblock \emph{arXiv preprint arXiv:2207.01780}.

\bibitem[{Lewis et~al.(2019)Lewis, Liu, Goyal, Ghazvininejad, Mohamed, Levy,
  Stoyanov, and Zettlemoyer}]{lewis2019bart}
Mike Lewis, Yinhan Liu, Naman Goyal, Marjan Ghazvininejad, Abdelrahman Mohamed,
  Omer Levy, Ves Stoyanov, and Luke Zettlemoyer. 2019.
\newblock Bart: Denoising sequence-to-sequence pre-training for natural
  language generation, translation, and comprehension.
\newblock \emph{arXiv preprint arXiv:1910.13461}.

\bibitem[{Li et~al.(2022{\natexlab{a}})Li, Hou, and Che}]{LI202271}
Bohan Li, Yutai Hou, and Wanxiang Che. 2022{\natexlab{a}}.
\newblock \href {https://doi.org/https://doi.org/10.1016/j.aiopen.2022.03.001}
  {Data augmentation approaches in natural language processing: A survey}.
\newblock \emph{AI Open}, 3:71--90.

\bibitem[{Li et~al.(2022{\natexlab{b}})Li, Choi, Chung, Kushman, Schrittwieser,
  Leblond, Eccles, Keeling, Gimeno, Lago, Hubert, Choy, de~Masson~d'Autume,
  Babuschkin, Chen, Huang, Welbl, Gowal, Cherepanov, Molloy, Mankowitz, Robson,
  Kohli, de~Freitas, Kavukcuoglu, and Vinyals}]{li2022competitionlevel}
Yujia Li, David Choi, Junyoung Chung, Nate Kushman, Julian Schrittwieser, Rémi
  Leblond, Tom Eccles, James Keeling, Felix Gimeno, Agustin~Dal Lago, Thomas
  Hubert, Peter Choy, Cyprien de~Masson~d'Autume, Igor Babuschkin, Xinyun Chen,
  Po-Sen Huang, Johannes Welbl, Sven Gowal, Alexey Cherepanov, James Molloy,
  Daniel~J. Mankowitz, Esme~Sutherland Robson, Pushmeet Kohli, Nando
  de~Freitas, Koray Kavukcuoglu, and Oriol Vinyals. 2022{\natexlab{b}}.
\newblock \href {http://arxiv.org/abs/2203.07814} {Competition-level code
  generation with alphacode}.

\bibitem[{Lin et~al.(2017)Lin, Li, He, Zhang, and Sun}]{NIPS2017_bf201d54}
Kevin Lin, Dianqi Li, Xiaodong He, Zhengyou Zhang, and Ming-ting Sun. 2017.
\newblock \href
  {https://proceedings.neurips.cc/paper/2017/file/bf201d5407a6509fa536afc4b380577e-Paper.pdf}
  {Adversarial ranking for language generation}.
\newblock In \emph{Advances in Neural Information Processing Systems},
  volume~30. Curran Associates, Inc.

\bibitem[{Liu et~al.(2020)Liu, Yan, Gong, Qi, Zhang, Jiao, Chen, Fu, Shou, Gong
  et~al.}]{liu2020glge}
Dayiheng Liu, Yu~Yan, Yeyun Gong, Weizhen Qi, Hang Zhang, Jian Jiao, Weizhu
  Chen, Jie Fu, Linjun Shou, Ming Gong, et~al. 2020.
\newblock Glge: A new general language generation evaluation benchmark.
\newblock \emph{arXiv preprint arXiv:2011.11928}.

\bibitem[{{Liu} et~al.(2019){Liu}, {Ott}, {Goyal}, {Du}, {Joshi}, {Chen},
  {Levy}, {Lewis}, {Zettlemoyer}, and {Stoyanov}}]{liu2019roberta}
Yinhan {Liu}, Myle {Ott}, Naman {Goyal}, Jingfei {Du}, Mandar {Joshi}, Danqi
  {Chen}, Omer {Levy}, Mike {Lewis}, Luke {Zettlemoyer}, and Veselin
  {Stoyanov}. 2019.
\newblock Roberta: A robustly optimized bert pretraining approach.
\newblock \emph{arXiv preprint arXiv:1907.11692}.

\bibitem[{Liu et~al.(2021)Liu, Dou, and Liu}]{liu-etal-2021-refsum}
Yixin Liu, Zi-Yi Dou, and Pengfei Liu. 2021.
\newblock \href {https://doi.org/10.18653/v1/2021.naacl-main.113} {{R}ef{S}um:
  Refactoring neural summarization}.
\newblock In \emph{Proceedings of the 2021 Conference of the North American
  Chapter of the Association for Computational Linguistics: Human Language
  Technologies}, pages 1437--1448, Online. Association for Computational
  Linguistics.

\bibitem[{Liu and Liu(2021)}]{liu-liu-2021-simcls}
Yixin Liu and Pengfei Liu. 2021.
\newblock \href {https://doi.org/10.18653/v1/2021.acl-short.135} {{S}im{CLS}: A
  simple framework for contrastive learning of abstractive summarization}.
\newblock In \emph{Proceedings of the 59th Annual Meeting of the Association
  for Computational Linguistics and the 11th International Joint Conference on
  Natural Language Processing (Volume 2: Short Papers)}, pages 1065--1072,
  Online. Association for Computational Linguistics.

\bibitem[{Liu et~al.(2022)Liu, Liu, Radev, and Neubig}]{liu-etal-2022-brio}
Yixin Liu, Pengfei Liu, Dragomir Radev, and Graham Neubig. 2022.
\newblock \href {https://doi.org/10.18653/v1/2022.acl-long.207} {{BRIO}:
  Bringing order to abstractive summarization}.
\newblock In \emph{Proceedings of the 60th Annual Meeting of the Association
  for Computational Linguistics (Volume 1: Long Papers)}, pages 2890--2903,
  Dublin, Ireland. Association for Computational Linguistics.

\bibitem[{Meister et~al.(2020)Meister, Cotterell, and
  Vieira}]{meister-etal-2020-beam}
Clara Meister, Ryan Cotterell, and Tim Vieira. 2020.
\newblock \href {https://doi.org/10.18653/v1/2020.emnlp-main.170} {If beam
  search is the answer, what was the question?}
\newblock In \emph{Proceedings of the 2020 Conference on Empirical Methods in
  Natural Language Processing (EMNLP)}, pages 2173--2185, Online. Association
  for Computational Linguistics.

\bibitem[{Pang et~al.(2021)Pang, He, and Cho}]{pang2021amortized}
Richard~Yuanzhe Pang, He~He, and Kyunghyun Cho. 2021.
\newblock Amortized noisy channel neural machine translation.
\newblock \emph{arXiv preprint arXiv:2112.08670}.

\bibitem[{Paulus et~al.(2018)Paulus, Xiong, and Socher}]{paulus2018a}
Romain Paulus, Caiming Xiong, and Richard Socher. 2018.
\newblock \href {https://openreview.net/forum?id=HkAClQgA-} {A deep reinforced
  model for abstractive summarization}.
\newblock In \emph{International Conference on Learning Representations}.

\bibitem[{Qi et~al.(2021)Qi, Gong, Yan, Xu, Yao, Zhou, Cheng, Jiang, Chen,
  Zhang et~al.}]{qi2021prophetnet}
Weizhen Qi, Yeyun Gong, Yu~Yan, Can Xu, Bolun Yao, Bartuer Zhou, Biao Cheng,
  Daxin Jiang, Jiusheng Chen, Ruofei Zhang, et~al. 2021.
\newblock Prophetnet-x: large-scale pre-training models for english, chinese,
  multi-lingual, dialog, and code generation.
\newblock \emph{arXiv preprint arXiv:2104.08006}.

\bibitem[{Qi et~al.(2020)Qi, Yan, Gong, Liu, Duan, Chen, Zhang, and
  Zhou}]{qi2020prophetnet}
Weizhen Qi, Yu~Yan, Yeyun Gong, Dayiheng Liu, Nan Duan, Jiusheng Chen, Ruofei
  Zhang, and Ming Zhou. 2020.
\newblock Prophetnet: Predicting future n-gram for sequence-to-sequence
  pre-training.
\newblock \emph{arXiv preprint arXiv:2001.04063}.

\bibitem[{Radford et~al.(2019)Radford, Wu, Child, Luan, Amodei, and
  Sutskever}]{radford2019language}
Alec Radford, Jeff Wu, Rewon Child, David Luan, Dario Amodei, and Ilya
  Sutskever. 2019.
\newblock Language models are unsupervised multitask learners.

\bibitem[{Raffel et~al.(2020)Raffel, Shazeer, Roberts, Lee, Narang, Matena,
  Zhou, Li, and Liu}]{JMLR:v21:20-074}
Colin Raffel, Noam Shazeer, Adam Roberts, Katherine Lee, Sharan Narang, Michael
  Matena, Yanqi Zhou, Wei Li, and Peter~J. Liu. 2020.
\newblock \href {http://jmlr.org/papers/v21/20-074.html} {Exploring the limits
  of transfer learning with a unified text-to-text transformer}.
\newblock \emph{Journal of Machine Learning Research}, 21(140):1--67.

\bibitem[{Rajpurkar et~al.(2016)Rajpurkar, Zhang, Lopyrev, and
  Liang}]{rajpurkar2016squad}
Pranav Rajpurkar, Jian Zhang, Konstantin Lopyrev, and Percy Liang. 2016.
\newblock \href {https://doi.org/10.18653/v1/D16-1264} {{SQ}u{AD}: 100,000+
  questions for machine comprehension of text}.
\newblock In \emph{Proceedings of the 2016 Conference on Empirical Methods in
  Natural Language Processing}, pages 2383--2392, Austin, Texas. Association
  for Computational Linguistics.

\bibitem[{Ravaut et~al.(2022)Ravaut, Joty, and Chen}]{ravaut2022summareranker}
Mathieu Ravaut, Shafiq Joty, and Nancy~F Chen. 2022.
\newblock Summareranker: A multi-task mixture-of-experts re-ranking framework
  for abstractive summarization.
\newblock \emph{arXiv preprint arXiv:2203.06569}.

\bibitem[{Ren et~al.(2021)Ren, Qu, Liu, Zhao, She, Wu, Wang, and
  Wen}]{ren2021rocketqav2}
Ruiyang Ren, Yingqi Qu, Jing Liu, Wayne~Xin Zhao, Qiaoqiao She, Hua Wu, Haifeng
  Wang, and Ji-Rong Wen. 2021.
\newblock Rocketqav2: A joint training method for dense passage retrieval and
  passage re-ranking.
\newblock In \emph{Proceedings of the 2021 Conference on Empirical Methods in
  Natural Language Processing}, pages 2825--2835.

\bibitem[{Rennie et~al.(2017)Rennie, Marcheret, Mroueh, Ross, and
  Goel}]{8099614}
Steven~J. Rennie, Etienne Marcheret, Youssef Mroueh, Jerret Ross, and Vaibhava
  Goel. 2017.
\newblock \href {https://doi.org/10.1109/CVPR.2017.131} {Self-critical sequence
  training for image captioning}.
\newblock In \emph{2017 IEEE Conference on Computer Vision and Pattern
  Recognition (CVPR)}, pages 1179--1195.

\bibitem[{Roller et~al.(2021)Roller, Dinan, Goyal, Ju, Williamson, Liu, Xu,
  Ott, Smith, Boureau, and Weston}]{roller-etal-2021-recipes}
Stephen Roller, Emily Dinan, Naman Goyal, Da~Ju, Mary Williamson, Yinhan Liu,
  Jing Xu, Myle Ott, Eric~Michael Smith, Y-Lan Boureau, and Jason Weston. 2021.
\newblock \href {https://doi.org/10.18653/v1/2021.eacl-main.24} {Recipes for
  building an open-domain chatbot}.
\newblock In \emph{Proceedings of the 16th Conference of the European Chapter
  of the Association for Computational Linguistics: Main Volume}, pages
  300--325, Online. Association for Computational Linguistics.

\bibitem[{Scialom et~al.(2021{\natexlab{a}})Scialom, Dray, Lamprier,
  Piwowarski, Staiano, Wang, and Gallinari}]{scialom-etal-2021-questeval}
Thomas Scialom, Paul-Alexis Dray, Sylvain Lamprier, Benjamin Piwowarski, Jacopo
  Staiano, Alex Wang, and Patrick Gallinari. 2021{\natexlab{a}}.
\newblock \href {https://doi.org/10.18653/v1/2021.emnlp-main.529}
  {{Q}uest{E}val: Summarization asks for fact-based evaluation}.
\newblock In \emph{Proceedings of the 2021 Conference on Empirical Methods in
  Natural Language Processing}, pages 6594--6604, Online and Punta Cana,
  Dominican Republic. Association for Computational Linguistics.

\bibitem[{Scialom et~al.(2021{\natexlab{b}})Scialom, Dray, Staiano, Lamprier,
  and Piwowarski}]{scialom2021beam}
Thomas Scialom, Paul-Alexis Dray, Jacopo Staiano, Sylvain Lamprier, and
  Benjamin Piwowarski. 2021{\natexlab{b}}.
\newblock To beam or not to beam: That is a question of cooperation for
  language gans.
\newblock \emph{Advances in neural information processing systems},
  34:26585--26597.

\bibitem[{See et~al.(2017)See, Liu, and Manning}]{see2017get}
Abigail See, Peter~J Liu, and Christopher~D Manning. 2017.
\newblock Get to the point: Summarization with pointer-generator networks.
\newblock \emph{arXiv preprint arXiv:1704.04368}.

\bibitem[{Shen et~al.(2015)Shen, Cheng, He, He, Wu, Sun, and
  Liu}]{DBLP:journals/corr/ShenCHHWSL15}
Shiqi Shen, Yong Cheng, Zhongjun He, Wei He, Hua Wu, Maosong Sun, and Yang Liu.
  2015.
\newblock \href {http://arxiv.org/abs/1512.02433} {Minimum risk training for
  neural machine translation}.
\newblock \emph{CoRR}, abs/1512.02433.

\bibitem[{Song et~al.(2019)Song, Tan, Qin, Lu, and Liu}]{pmlr-v97-song19d}
Kaitao Song, Xu~Tan, Tao Qin, Jianfeng Lu, and Tie-Yan Liu. 2019.
\newblock \href {https://proceedings.mlr.press/v97/song19d.html} {{MASS}:
  Masked sequence to sequence pre-training for language generation}.
\newblock In \emph{Proceedings of the 36th International Conference on Machine
  Learning}, volume~97 of \emph{Proceedings of Machine Learning Research},
  pages 5926--5936. PMLR.

\bibitem[{Sutton et~al.(1999)Sutton, McAllester, Singh, and
  Mansour}]{NIPS1999_464d828b}
Richard~S Sutton, David McAllester, Satinder Singh, and Yishay Mansour. 1999.
\newblock \href
  {https://proceedings.neurips.cc/paper/1999/file/464d828b85b0bed98e80ade0a5c43b0f-Paper.pdf}
  {Policy gradient methods for reinforcement learning with function
  approximation}.
\newblock In \emph{Advances in Neural Information Processing Systems},
  volume~12. MIT Press.

\bibitem[{Vaswani et~al.(2017)Vaswani, Shazeer, Parmar, Uszkoreit, Jones,
  Gomez, Kaiser, and Polosukhin}]{vaswani2017attention}
Ashish Vaswani, Noam Shazeer, Niki Parmar, Jakob Uszkoreit, Llion Jones,
  Aidan~N Gomez, {\L}ukasz Kaiser, and Illia Polosukhin. 2017.
\newblock Attention is all you need.
\newblock \emph{Advances in neural information processing systems}, 30.

\bibitem[{Vijayakumar et~al.(2016)Vijayakumar, Cogswell, Selvaraju, Sun, Lee,
  Crandall, and Batra}]{diverse-beam}
Ashwin~K. Vijayakumar, Michael Cogswell, Ramprasaath~R. Selvaraju, Qing Sun,
  Stefan Lee, David~J. Crandall, and Dhruv Batra. 2016.
\newblock \href {http://arxiv.org/abs/1610.02424} {Diverse beam search:
  Decoding diverse solutions from neural sequence models}.
\newblock \emph{CoRR}, abs/1610.02424.

\bibitem[{Wen et~al.(2015)Wen, Gasic, Mrksic, Su, Vandyke, and
  Young}]{wen2015semantically}
Tsung-Hsien Wen, Milica Gasic, Nikola Mrksic, Pei-Hao Su, David Vandyke, and
  Steve Young. 2015.
\newblock Semantically conditioned lstm-based natural language generation for
  spoken dialogue systems.
\newblock \emph{arXiv preprint arXiv:1508.01745}.

\bibitem[{Wolf et~al.(2020)Wolf, Debut, Sanh, Chaumond, Delangue, Moi, Cistac,
  Rault, Louf, Funtowicz, Davison, Shleifer, von Platen, Ma, Jernite, Plu, Xu,
  Scao, Gugger, Drame, Lhoest, and Rush}]{wolf-etal-2020-transformers}
Thomas Wolf, Lysandre Debut, Victor Sanh, Julien Chaumond, Clement Delangue,
  Anthony Moi, Pierric Cistac, Tim Rault, Rémi Louf, Morgan Funtowicz, Joe
  Davison, Sam Shleifer, Patrick von Platen, Clara Ma, Yacine Jernite, Julien
  Plu, Canwen Xu, Teven~Le Scao, Sylvain Gugger, Mariama Drame, Quentin Lhoest,
  and Alexander~M. Rush. 2020.
\newblock \href {https://www.aclweb.org/anthology/2020.emnlp-demos.6}
  {Transformers: State-of-the-art natural language processing}.
\newblock In \emph{Proceedings of the 2020 Conference on Empirical Methods in
  Natural Language Processing: System Demonstrations}, pages 38--45, Online.
  Association for Computational Linguistics.

\bibitem[{Yu et~al.(2017)Yu, Zhang, Wang, and Yu}]{Yu_Zhang_Wang_Yu_2017}
Lantao Yu, Weinan Zhang, Jun Wang, and Yong Yu. 2017.
\newblock \href {https://doi.org/10.1609/aaai.v31i1.10804} {Seqgan: Sequence
  generative adversarial nets with policy gradient}.
\newblock \emph{Proceedings of the AAAI Conference on Artificial Intelligence},
  31(1).

\bibitem[{Zhang et~al.(2021)Zhang, Gong, Shen, Lv, Duan, and
  Chen}]{zhang2021adversarial}
Hang Zhang, Yeyun Gong, Yelong Shen, Jiancheng Lv, Nan Duan, and Weizhu Chen.
  2021.
\newblock Adversarial retriever-ranker for dense text retrieval.
\newblock \emph{arXiv preprint arXiv:2110.03611}.

\bibitem[{Zhang et~al.(2020)Zhang, Zhao, Saleh, and Liu}]{pmlr-v119-zhang20ae}
Jingqing Zhang, Yao Zhao, Mohammad Saleh, and Peter Liu. 2020.
\newblock \href {https://proceedings.mlr.press/v119/zhang20ae.html} {{PEGASUS}:
  Pre-training with extracted gap-sentences for abstractive summarization}.
\newblock In \emph{Proceedings of the 37th International Conference on Machine
  Learning}, volume 119 of \emph{Proceedings of Machine Learning Research},
  pages 11328--11339. PMLR.

\bibitem[{Zhang et~al.(2018)Zhang, Dinan, Urbanek, Szlam, Kiela, and
  Weston}]{zhang2018personalizing}
Saizheng Zhang, Emily Dinan, Jack Urbanek, Arthur Szlam, Douwe Kiela, and Jason
  Weston. 2018.
\newblock \href {https://doi.org/10.18653/v1/P18-1205} {Personalizing dialogue
  agents: {I} have a dog, do you have pets too?}
\newblock In \emph{Proceedings of the 56th Annual Meeting of the Association
  for Computational Linguistics (Volume 1: Long Papers)}, pages 2204--2213,
  Melbourne, Australia. Association for Computational Linguistics.

\bibitem[{Zhang* et~al.(2020)Zhang*, Kishore*, Wu*, Weinberger, and
  Artzi}]{Zhang*2020BERTScore:}
Tianyi Zhang*, Varsha Kishore*, Felix Wu*, Kilian~Q. Weinberger, and Yoav
  Artzi. 2020.
\newblock \href {https://openreview.net/forum?id=SkeHuCVFDr} {Bertscore:
  Evaluating text generation with bert}.
\newblock In \emph{International Conference on Learning Representations}.

\bibitem[{Zhao et~al.(2018)Zhao, Ni, Ding, and Ke}]{zhao2018paragraph}
Yao Zhao, Xiaochuan Ni, Yuanyuan Ding, and Qifa Ke. 2018.
\newblock \href {https://doi.org/10.18653/v1/D18-1424} {Paragraph-level neural
  question generation with maxout pointer and gated self-attention networks}.
\newblock In \emph{Proceedings of the 2018 Conference on Empirical Methods in
  Natural Language Processing}, pages 3901--3910, Brussels, Belgium.
  Association for Computational Linguistics.

\bibitem[{Zhong et~al.(2020)Zhong, Liu, Chen, Wang, Qiu, and
  Huang}]{zhong-etal-2020-extractive}
Ming Zhong, Pengfei Liu, Yiran Chen, Danqing Wang, Xipeng Qiu, and Xuanjing
  Huang. 2020.
\newblock \href {https://doi.org/10.18653/v1/2020.acl-main.552} {Extractive
  summarization as text matching}.
\newblock In \emph{Proceedings of the 58th Annual Meeting of the Association
  for Computational Linguistics}, pages 6197--6208, Online. Association for
  Computational Linguistics.

\end{thebibliography}
\bibliographystyle{acl_natbib}

\appendix

\section{Discussion}\label{app:discussion}
In this section, we discuss the relations between our JGR and several popular methods, including reinforcement learning (RL), generative adversarial networks (GAN), and data augmentation.

\subsection{JGR \& RL}\label{app:RL}
Some previous RL works, i.e., ~\citep{DBLP:journals/corr/ShenCHHWSL15, 8099614, paulus2018a} proposed to use $\Delta(\mathbf{\hat{y}}, \mathbf{y})$ to compute reward $\mathcal{R}(\mathbf{\hat{y}})$ directly which doesn't combine ranking scores as feedback signals. However, we argue that the ranking score calculated by the ranker model can provide more semantic-relevant information than the matching scores (e.g., BLEU or ROUGE) which are purely based on the surface match. In the ablation study, we also demonstrate that the proposed approach is superior to other configurations in terms of training stability and performance. 

Some other RL works~\citet{bahdanau2017an, le2022coderl} introduced actor-critic frameworks~\citep{konda1999actor}, which jointly train an actor and a critic, are similar to our JGR framework. 
However, they have not considered the contrastive rewards between different candidates given one input.
Different from these works, JGR allows the generator to sample several i.i.d. candidates and be optimized simultaneously on these candidates at each training step. This improvement makes the reward of a sampled candidate contain contrastive information from the candidates from the same candidate set. Furthermore, it effectively raises the number of diverse chains of probabilities on which the generator can be optimized. In Table \ref{tab:res_reward}, we compare our JGR-G with the simple self-critical that uses metric rewards, and the actor-critic baseline that the critic is trained to fit the metric score $\Delta(\mathbf{\hat{y}}, \mathbf{y})$. The empirical results show that trained with the JGR framework, the generator model can surpass those trained with previous RL-based methods well used in the NLG area.

\subsection{JGR \& GAN}\label{app:gan}
From the perspective of the composition of a framework, both JGR and GAN contain a generator and a critic. In GAN, the critic is the discriminator that aims at discriminating the real candidate from the candidate pool. While in JGR, the critic is the ranker that aims to re-rank the candidates generated by the generator.

The main difference between JGR and GAN comes from the training objective.  Let the $G_{\theta}$ denotes the generator, and $D_{\phi}$ denotes the discriminator/ranker. GAN trains $G_{\theta}$ and $D_{\phi}$ with the min-max objective:
\begin{equation}
\label{eq:gan_obj}
\begin{split}
    \mathcal{J}_{G_{\theta} D_{\phi}} = \text{min}_{\theta} \text{max}_{\phi} E_{\mathbf{y}^+ \sim p_{\text{true}}(\cdot|\mathbf{x}) }[\text{log}p_{D_{\phi}}(\mathbf{y}^+,\textbf{x})] \\ + E_{\hat{\mathbf{y}}^- \sim p_{G_{\theta}}(\cdot|\mathbf{x})}[\text{log}(1-p_{D_{\phi}}(\hat{\mathbf{y}}^-,\textbf{x}))]
\end{split}
\end{equation}

In Eq.~\ref{eq:gan_obj}, $\mathbf{y}^+$ is the ground-truth output of input $\mathbf{x}$, and $\hat{\mathbf{y}}^-$ is the candidate texts sampled by the generator. This is different from the setting of JGR, where both $\mathbf{y}^+$ (denoted as $\hat{\mathbf{y}}^+$ in JGR) and $\hat{\mathbf{y}}^-$ are sampled from $p_{G_{\theta}}(\cdot|\mathbf{x})$. 

To implement GAN in NLG, according to \cite{Yu_Zhang_Wang_Yu_2017}, the policy gradient is used and the reward assigned to $\hat{\mathbf{y}}^-$ is $\text{log}p_{D_{\phi}}(\hat{\mathbf{y}}^-,\textbf{x})$. Note that the reward is always positive, therefore GAN essentially raises the probability of the generator outputs, regardless of the quality of the outputs. On contrary, as computed in Eq.~\ref{eq:loss_jgr}, there are both positive and negative rewards in JGR, which means that JGR not only encourages the generator to generate good candidates but also punishes the generator when generating bad candidates.

\begin{table}[h]
	\centering
    \resizebox{0.95\linewidth}{!}{
	\begin{tabular}{lcccccc}
		\toprule
		&R-1& R-2 & R-L& AVG\\
		\hline
		BART & 44.16& 21.28 &40.90 & \cellcolor[HTML]{ECF4FF}35.45\\
		\hline
		GAN$_{\text{std}}$&43.68 &20.81 &40.45 &\cellcolor[HTML]{ECF4FF}34.98 \\
		GAN$_{\text{mod}}$&42.93 &20.66 &39.87 &\cellcolor[HTML]{ECF4FF}34.49 \\
		\hline
		JGR-G&\textbf{46.86} &\textbf{23.18} &\textbf{43.74} &\cellcolor[HTML]{ECF4FF}\textbf{37.93} \\
		\bottomrule
	\end{tabular}
	}
	\captionof{table}{Results generator in JGR and two kinds of GANs.}
	\label{tab:res_vs_gan}
\end{table}

Table~\ref{tab:res_vs_gan} shows the performance of generators in JGR and GAN on CNN/DailyMail, where GAN$_{\text{std}}$ is the standard GAN setting that $\mathbf{y}^+$ is the ground-truth text and GAN$_{\text{mod}}$ is our modified version of GAN that $\mathbf{y}^+$ is replaced by the best candidate sampled by the generator, i.e., $\hat{\mathbf{y}}^+$. As shown in the table, our JGR surpasses the GAN methods, and the performance of GAN$_{\text{std}}$ and GAN$_{\text{mod}}$ can not even surpass the model trained on optimizing the standard NLL loss, indicating that the GAN methods are not suitable for all NLG tasks. The GAN$_{\text{mod}}$ performs worse than GAN$_{\text{std}}$, showing that for the min-max objective of GAN, it is not a good choice to letting $\hat{\mathbf{y}}^+$ as the positive sample, which is contrary to what we found in JGR.

\subsection{JGR \& Data Augmentation}

Data augmentation methods aim to improve the models' performance by adding modified or synthesized data to the existing training data~\citep{LI202271}. For natural language generation tasks, denote the augmented dataset as $\hat{\mathcal{D}}$, where $\hat{\mathcal{D}}$ contains several augmented samples $(\hat{\mathbf{x}}, \hat{\mathbf{y}})$, the training object for model in the augmented data is:
\begin{equation}\label{eq:loss_da}
    \mathcal{L}_{\text{DA}} = -\sum\limits_{\mathbf{(\hat{\mathbf{x}}, \hat{\mathbf{y}})}\in\hat{\mathcal{D}}} \sum\limits_{t}\log p_{G_{\theta}}(\hat{y}_t|\hat{y}_{<t},\mathbf{\hat{x}})
\end{equation}
The above equation is similar to JGR's reinforcement learning loss in Eq.~\ref{eq:loss_jgr}. Both of them optimize the generator by maximizing the log-likelihood of synthesized data.  Therefore, from this perspective, we can regard our JGR as a way of data augmentation where the synthesized data is sampled from the generator and the log-likelihood is re-scaled by the rewards.  

\begin{table}[h]
	\centering
    \resizebox{0.95\linewidth}{!}{
	\begin{tabular}{lcccccc}
		\toprule
		&R-1& R-2 & R-L& AVG\\
		\hline
		BART & 44.16& 21.28 &40.90 & \cellcolor[HTML]{ECF4FF}35.45\\
		\hline
		DA$_\text{sep}$&44.37 &21.24 &41.18 &\cellcolor[HTML]{ECF4FF}35.60 \\
		DA$_\text{mix}$&44.27 &21.38 &41.04 &\cellcolor[HTML]{ECF4FF}35.56 \\
		\hline
		JGR-G&\textbf{46.86} &\textbf{23.18} &\textbf{43.74} &\cellcolor[HTML]{ECF4FF}\textbf{37.93} \\
		\bottomrule
	\end{tabular}
	}
	\captionof{table}{Results generator in JGR and two kinds of GANs.}
	\label{tab:res_vs_da}
	
\end{table}

We designed two simple but effective data augmentation methods named DA$_\text{sep}$ and DA$_\text{mix}$. Both of DA$_\text{sep}$ and DA$_\text{mix}$ use a fine-tuned generator $G^{0}$ to generate one summary $\hat{\mathbf{y}}$ for each input $\mathbf{x}$ in original training set $\mathcal{D}$ using beam search, the collection of all $(\mathbf{x},\hat{\mathbf{y}})$ is treat as the augmented training data $\mathcal{\hat{D}}$. After that, 1) DA$_\text{sep}$ fine-tunes $G^{0}$ firstly on $\mathcal{\hat{D}}$ and then on $\mathcal{D}$, 2) DA$_\text{mix}$ further fine-tunes $G^{0}$ on the mixture of $\mathcal{\hat{D}}$ and $\mathcal{D}$. We compare the performance of DA$_\text{sep}$ and DA$_\text{mix}$ with our JGR on CNN/DailyMail, with BART as the generator, the results are shown in Table~\ref{tab:res_vs_da}. As shown in the results, both DA$_\text{sep}$ and DA$_\text{mix}$ can further improve the performance of BART, verifying the effect of data augmentation. However, the performance gain brought by data augmentation is far less than that brought by JGR.


\section{Computation of Self-BLEU}\label{app:self_bleu}

Given a candidate set $\mathcal{\hat{Y}} = \{ \mathbf{\hat{y}}^1,\mathbf{\hat{y}}^2, ..., \mathbf{\hat{y}}^C \}$ sampled from the generator, the self-BLEU score for $\mathcal{\hat{Y}}$ is computed as the average of mutual BLEU scores of all candidate pairs:
\begin{equation}
    \text{self-BLEU}(\mathcal{\hat{Y}}) = \frac{\sum\limits_{\mathbf{\hat{y}}^i,\mathbf{\hat{y}}^j\in\mathcal{\hat{Y}};i\neq j}\text{BLEU}(\mathbf{\hat{y}}^i,\mathbf{\hat{y}}^j)}{C(C-1)}
\end{equation}
A higher self-BLEU score means the sampled candidates are more similar to each other, in other words, a lower diversity of the sampled candidates.

\begin{figure}[h]
    \centering
    \includegraphics[width=0.9\linewidth]{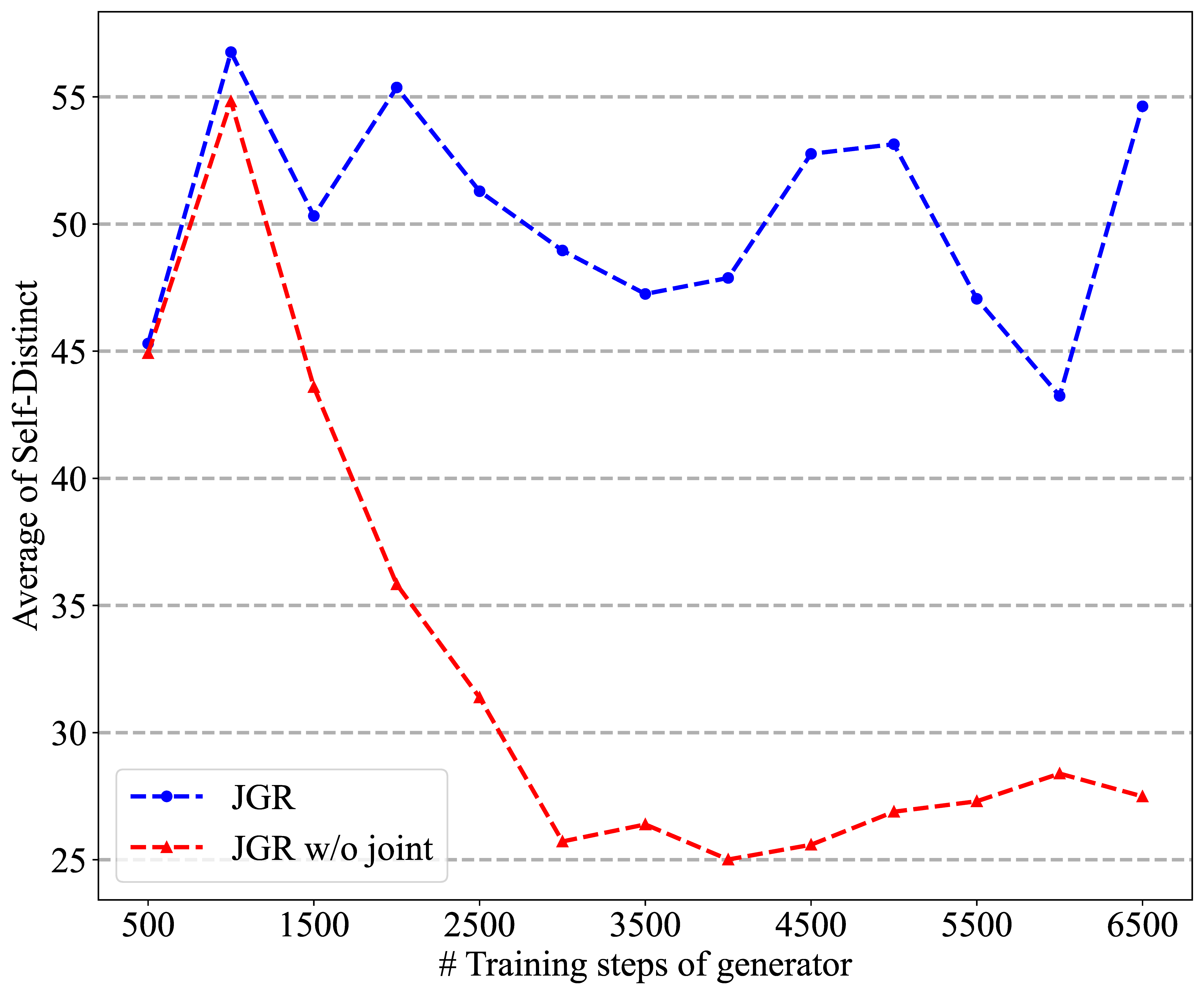}
	\caption{The average of self-Distinct-2 at each training interval.  }
	\label{fig:self-distinct}
	\vspace{-0.3cm}
\end{figure}

It is another way to assess the diversity of sampled candidates by computing the proportion of the number of distinct n-grams in the total number of tokens for the sampled candidates of an input sequence. We refer to this metric as self-Distinct-n where n refers to n-grams. The higher self-Distinct-n corresponds to the higher diversity of sampled candidates. Like Figure \ref{fig:std-rewards}, we show the curves of the average self-Distinct-2 for generated candidates at each training interval in Figure \ref{fig:self-distinct}. From the figure, we can see that the self-Distinct-2 of JGR$_{\text{w/o joint}}$ drops rapidly after the generator is trained 1000 steps, while the self-Distinct-2 keeps hovering in a relatively high range for JGR. This phenomenon aligns with what we found when applying self-BLEU and further enhances our conclusion in Section \ref{sec:joint_train}.

\section{Decoding Strategies}\label{app:decode}

\begin{figure}[h]
    \centering
    \includegraphics[width=0.95\linewidth]{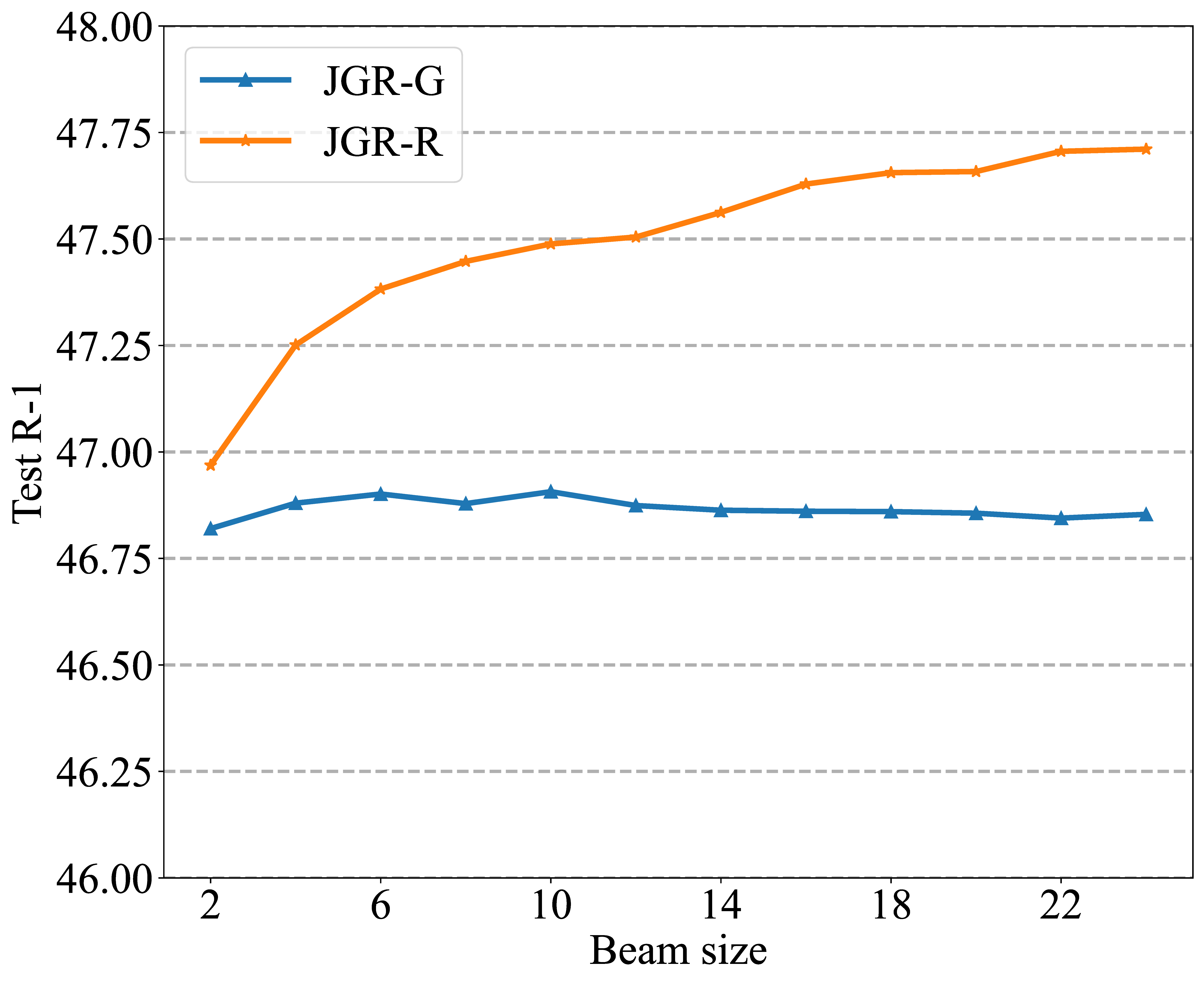}
	\caption{ROUGE-1 score using beam search with different on CNN/DailyMail test set. }
	\label{fig:ncand_beam}
	\vspace{-0.3cm}
\end{figure}

We study the impact of different decoding strategies during inference. Two decoding strategies are compared, namely beam search and group beam search~\citep{diverse-beam}. We also compare different beam sizes. The results of ROUGE-1 score with beam search on CNN/DailyMail are shown in Figure \ref{fig:ncand_beam}.

\begin{figure}[h]
    \centering
    \includegraphics[width=0.95\linewidth]{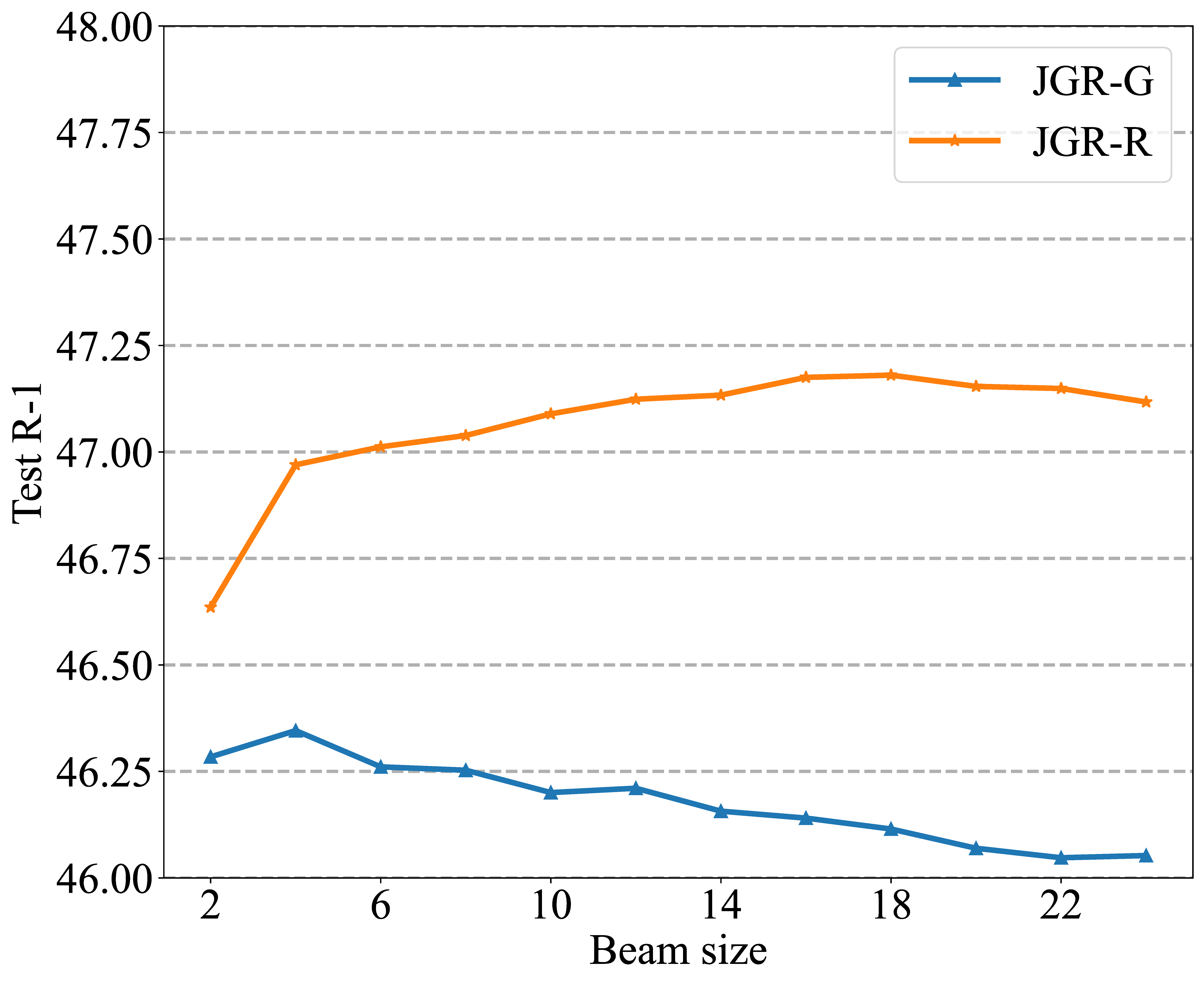}
	\caption{ROUGE-1 score using diverse beam search with different on CNN/DailyMail test set. }
    	\label{fig:ncand_group}
	\vspace{-0.3cm}
\end{figure}


As shown in Figure \ref{fig:ncand_beam}, increasing the beam size does not contribute to the performance of JGR-G when using the normal beam search. However, the performance of JGR-R can rise as the beam size increases. This indicates that increasing the beam size can raise the probability of JGR-R ranking a better candidate to the top among all the candidates decoded by JGR-G.


Figure \ref{fig:ncand_group} shows the results with diverse beam search. Firstly we can find that with diverse beam search the JGR system can not achieve comparable results with JGR using normal beam search, and the performance of JGR-G begins to drop when beam size exceeds 4. We can still observe that the performance of JGR-R rises as the beam size increases. However, since the performance of JGR-G keeps declining, the performance ascent of JGR-R is not as significant as that of JGR-R with the normal beam search.

\section{Details of Human Evaluation}\label{app:human-eval}

We conduct a human evaluation on CNN/DailyMail. Following Blenerbot v2~\citep{roller-etal-2021-recipes}, we ask the annotators to explicitly compare which generated text is better for each pair of generated outputs, rather than assign an evaluation score. This explicit comparison can avoid the per annotator bias in numerical scores (e.g., annotators who tend to give generous scores), and remedy many of the issues of sequential effects such as contrasting with a previous case. We randomly picked 100 cases from the CNN/DailyMail test set, each case was organized as <Doc, Summary \#1, Summary \#2> where Doc means the source document, Summary \#1 and Summary  \#2 mean the summaries generated by JGR and BART.

The annotators were asked to compare Summary \#1 and Summary \#2 on three aspects given at the end of each case. To avoid the stereotype of annotators that Summary \#1 or Summary \#2 is better according to previous cases, we randomly shuffle the summaries in each case, which means that Summary \#1 is not necessarily from JGR or BART, and so as Summary \#2.

Each picked case was annotated by 3 annotators, and they worked individually without communication. Given a certain human evaluation metric on one case, the comparison result is obtained by the following rules:

\begin{itemize}
    \item If more than or equal to two annotators think JGR has won in that metric, then JGR wins.
    \item If more than or equal to two annotators think BART has won in that metric, then BART wins. 
    \item Otherwise, the comparison result is marked as a tie.
\end{itemize}

We evaluate JGR and BART from three aspects, namely informativeness (Inform.), factual consistency (Fact.), and readability (Read.). The results are shown in Table~\ref{tab:res_human}. Note that since we use direct comparison, the number of “tie” cases may be fewer than some works that conduct human evaluation through assigning scores. 

\section{Details of JGR$_{\text{w/o joint}}$}\label{app:without-joint}

To implement JGR$_{\text{w/o joint}}$, we first fully train the generator with the negative likelihood loss. Then we use this generator to generate candidates and fully train the ranker with the objective described in Eq.~\ref{eq:ranker_obj}. Then we train the generator again using the same RL paradigm as JGR with the reward from the ranker. The only different between  JGR$_{\text{w/o joint}}$ and JGR is that JGR$_{\text{w/o joint}}$ does not incorporate the iterative training.

\section{Details of the Benchmarks and Evaluation Metrics}\label{sec:statistic}

\noindent\textbf{CNN/DailyMail}~\citep{hermann2015cnndm} is a benchmark for summarization. Both extractive and abstractive summarization models can be applied on CNN/DailyMail. Since our JGR focuses on text generation, we treat CNN/DailyMail as an abstractive summarization task. There are two versions: anonymized and non-anonymized. We use the non-anonymized dataset~\citet{see2017get}. The evaluation metrics are Rouge-1, Rouge-2, and Rouge-L. 

\noindent\textbf{SAMSum}~\citep{gliwa2019samsum} is a benchmark for conversational summarization, whose inputs are the concatenation of dialog context. The evaluation metrics are Rouge-1, Rouge-2, and Rouge-L. 

\noindent\textbf{SQuAD 1.1}~\citep{rajpurkar2016squad} is originally an machine reading comprehension dataset. We follow the data split and pre-processing as done by ~\citet{du2017learning,zhao2018paragraph,liu2020glge}, to make it a question generation dataset, which treats the concatenation of the answer span and article as the input, and the question as the target output. The evaluation metrics are Rouge-L, Bleu-4, and METEOR.

\noindent\textbf{PersonaChat}~\cite{zhang2018personalizing} contains about 160K utterances. Given the multi-turn conversations and persona profile, the model learns to generate the response. The evaluation metrics are Bleu-1, Bleu-2, and the ratio of distinct unigrams and bigrams in the generated responses (Distinct-1 and Distinct-2).

The statistics of all benchmarks are shown in Table~\ref{tab:statistic}.

\begin{table}[h]
	\centering
	\resizebox{0.98\linewidth}{!}{
	\begin{tabular}{lrrrrr}
		\toprule
		Benchmark & $|\text{Train}|$ & $|\text{Dev}|$ & $|\text{Test}|$& $|\text{Src.}|$ & $|\text{Tgt.}|$\\
		\hline
		CNN/DailyMail& 287,113& 13,368& 11,490& 822.3& 57.9 \\
		SAMSum & 14,731 & 818 & 819 & 124.1 & 23.4 \\
		SQuAD 1.1 &75,722 & 10,570 & 11,877 & 149.4 & 11.5 \\
		PersonaChat & 122,499 & 14,602& 14,056& 120.8 & 11.8 \\
		\bottomrule
	\end{tabular}
	}
	\caption{The statistics of the benchmarks.  $|\text{Src.}|$ means the average number of tokens for each source input. $|\text{Tgt.}|$ means the average number of tokens for each target text.
}
	\label{tab:statistic}
\end{table}

For evaluation on CNN/Daily and SAMSum, we use the python rouge score package: \url{https://pypi.org/project/rouge-score/}. For evaluation on SQuAD 1.1, we follow the evaluation scripts open-sourced by \citet{liu2020glge} at \url{https://github.com/microsoft/ProphetNet/tree/master/GLGE_baselines/script/script/evaluate/qg}. For evaluation on PersonaChat, we follow the evaluation scripts open-sourced by \citet{liu2020glge} at \url{https://github.com/microsoft/ProphetNet/tree/master/GLGE_baselines/script/script/evaluate/personachat}.

\begin{table*}[t]
	\centering
	\resizebox{1.0\textwidth}{!}{
	\begin{tabular}{lcccc}
		\toprule
		 &\makebox[0.24\textwidth]{CNN/DailyMain} & \makebox[0.24\textwidth]{SAMSum} & \makebox[0.24\textwidth]{SQuAD 1.1} & \makebox[0.24\textwidth]{PersonaChat} \\
		\hline
		\multicolumn{5}{c}{Warming-up $G^0$ } \\
		\hline
		\# Epochs& 5  & 5& 20 & 5\\
		Learning rate& 5e-5 & 5e-5& 5e-5 & 5e-5\\
		Batch size& 96& 128& 96 & 96\\
		Max source length& 1024& 1024& 600 & 700\\
		Max target length& 100& 100& 65 & 70\\
		\hline
		\multicolumn{5}{c}{First Ranker training iteration } \\
		\hline
		\# Epochs& 3 &20 &3 &3 \\
		Learning rate& 1e-5 &1e-5 &1e-5&1e-5 \\
		Warm-up ratio/steps& 0.2 & 500 steps&0.2 & 0.3\\
		Batch size& 64  & 64&64 & 32\\
		Max source length& 512 &512 &500&500 \\
		\# Candidates sampled for $G^0$&\multicolumn{4}{c}{16} \\
		\# Negative candidates&\multicolumn{4}{c}{2}\\
		$\Delta(\mathbf{\hat{y}}, \mathbf{y})$ &\multicolumn{2}{c}{0.02(R-1)+0.05(R-2)+0.025(R-L)} &0.02(R-L)+0.04(B-4)+0.04(MTR) &0.02(B-1)+0.025(B-2)  \\
		\hline
		\multicolumn{5}{c}{JGR training } \\
		\hline
		\# Epochs& 3  & 10 &3 &3\\
		\# JGR-R steps per iteration&500 & 231 steps (1 epoch) &250 &500 \\
		\# JGR-G steps per iteration&500 &231 steps (1 epoch) &250 &500 \\
		JGR-G learning rate&5e-5  &1e-5 &5e-5 &5e-5 \\
		JGR-R learning rate&1e-5 & 5e-6&1e-5 &1e-5 \\
		Batch size&64  &64 &32 &64\\
		\# Candidates sampled for JGR-R&\multicolumn{4}{c}{16}  \\
		\# Negative candidates for JGR-R &\multicolumn{4}{c}{2}  \\
		\# Candidates sampled for JGR-G&\multicolumn{4}{c}{8}  \\
		Beam size when inference&\multicolumn{4}{c}{16} \\
		$\Delta(\mathbf{\hat{y}}, \mathbf{y})$&\multicolumn{2}{c}{0.02(R-1)+0.05(R-2)+0.025(R-L)} &0.02(R-L)+0.04(B-4)+0.04(MTR) &0.02(B-1)+0.025(B-2)  \\
		\bottomrule
	\end{tabular}
	}
	\caption{The hyper-parameters of JGR on each benchmark.
}
	\label{tab:hyper_parameters}
\end{table*}
\section{Hyper-parameters of Fine-tuning on Benchmarks.  }\label{sec:hyper_parameter}

The hyper-parameters for our JGR on each benchmark are shown in Table \ref{tab:hyper_parameters}.

\end{document}